\documentclass[sigconf]{acmart}

\usepackage{url}
\usepackage{amsmath}
\usepackage{extarrows}
\usepackage{multirow}
\usepackage{subfigure}
\usepackage{natbib}

\usepackage[noend]{algpseudocode}
\usepackage{algorithm}
\usepackage{algorithmicx}
\usepackage{xspace}
\usepackage{dsfont}
\usepackage{epsfig}
\usepackage{epstopdf}
\usepackage{bm}
\usepackage{color}
\usepackage{multirow}
\usepackage{booktabs} 
\usepackage{tablefootnote}
\usepackage{threeparttable} 
\usepackage{enumitem} 
\usepackage{makecell} 

\newtheorem{mydef}{Definition}

\newcommand{\paratitle}[1]{\vspace{1.1ex}\noindent\textbf{#1}}
\newcommand{\fig}{Fig.\xspace}
\newcommand{\tab}{Tab.\xspace}

\newcommand{\equ}{Eq.\xspace}

\newcommand{\ie}{\emph{i.e., }}

\newcommand{\eg}{\emph{e.g., }}

\newcommand{\etc}{\emph{etc}}
\newcommand{\wrt}{\emph{w.r.t. }}
\newcommand{\model}{STEVE\xspace}
\newcommand{\git}{\url{https://github.com/bigscity/STEVE_CODE}}

\AtBeginDocument{%
  }

\copyrightyear{2025} 
\acmYear{2025} 
\setcopyright{acmlicensed}
\acmConference[KDD '25]{Proceedings of the 31st ACM SIGKDD Conference on Knowledge Discovery and Data Mining V.1}{August 3--7, 2025}{Toronto, ON, Canada} 
\acmBooktitle{Proceedings of the 31st ACM SIGKDD Conference on Knowledge Discovery and Data Mining V.1 (KDD '25), August 3--7, 2025, Toronto, ON, Canada} 
\acmDOI{10.1145/3690624.3709201} 
\acmISBN{979-8-4007-1245-6/25/08}

\begin{document}

\title{Seeing the Unseen: Learning Basis Confounder Representations for Robust Traffic Prediction}

\author{Jiahao Ji}
\orcid{0000-0003-3029-2262}
\authornote{These authors contributed equally to this work.}
\author{Wentao Zhang}
\authornotemark[1]
\orcid{0009-0009-7285-6872}
\affiliation{%
  \institution{School of Computer\\Science and Engineering,\\Beihang University}
  \city{Beijing}
  \country{China}
}

\author{Jingyuan Wang}
\authornote{Corresponding author: jywang@buaa.edu.cn.}
\orcid{0000-0003-0651-1592}
\affiliation{%
  \institution{SCSE, Beihang University}
  \city{Beijing}
  \country{China}
}
\affiliation{
\institution{MIIT Key Laboratory of Data Intelligence and Management, SEM, Beihang University}
\city{Beijing}
  \country{China}
}

\author{Chao Huang}
\orcid{0000-0002-2062-1512}

\affiliation{%
  \institution{Department of Computer Science,\\Musketeers Foundation Institute\\of Data Science,\\University of Hong Kong}
  \city{Hong Kong SAR}
  \country{China}
}

\renewcommand{\shortauthors}{Jiahao Ji, Wentao Zhang, Jingyuan Wang, and Chao Huang}

\begin{abstract}
  Traffic prediction is essential for intelligent transportation systems and urban computing. It aims to establish a relationship between historical traffic data $X$ and future traffic states $Y$ by employing various statistical or deep learning methods. However, the relations of $X \rightarrow Y$ are often influenced by external confounders that simultaneously affect both $X$ and $Y$, such as weather, accidents, and holidays. Existing deep-learning traffic prediction models adopt the classic front-door and back-door adjustments to address the confounder issue.   However, these methods have limitations in addressing continuous or undefined confounders, as they depend on predefined discrete values that are often impractical in complex, real-world scenarios. To overcome this challenge, we propose the Spatial-Temporal sElf-superVised confoundEr learning (\model) model. This model introduces a basis vector approach, creating a base confounder bank to represent any confounder as a linear combination of a group of basis vectors. It also incorporates self-supervised auxiliary tasks to enhance the expressive power of the base confounder bank. Afterward, a confounder-irrelevant relation decoupling module is adopted to separate the confounder effects from direct $X \rightarrow Y$ relations. Extensive experiments across four large-scale datasets validate our model's superior performance in handling spatial and temporal distribution shifts and underscore its adaptability to unseen confounders. 
  Our model implementation is available at \git.
\end{abstract}



\maketitle

\section{Introduction}

Traffic prediction, a key technology in intelligent transportation systems and urban computing~\cite{libcity,chen2018price}, has long been a prominent research area in spatiotemporal data mining~\cite{ding2018ultraman,chen2019real}. A high-performance and robust traffic prediction model is crucial for efficient urban traffic management and safe city operations~\cite{ji2022stden}. 
Typically, it uses historical traffic states $X$ as inputs to predict future traffic states, denoted as $Y$, in upcoming time slots~\cite{stgcn,agcrn,ji2023spatio}. In the literature, numerous models have been proposed to capture the dependency relationship between $X$ and $Y$, including shallow statistical methods, such as ARIMA~\cite{arima}, SVR~\cite{svr}, and Kalman filtering~\cite{kalman_filter2014}, as well as deep learning-based methods in recent years. For instance, using recurrent neural networks~\cite{agcrn}, temporal convolutional networks~\cite{gwnet,ji2023spatio}, and transformers~\cite{pdformer2023} to model temporal correlations, as well as using convolutional neural networks~\cite{nycbike2} and graph neural networks~\cite{zhou2023maintain} to capture spatial dependencies.

\begin{figure}[t]
    \centering
    \includegraphics[width=0.85\columnwidth]{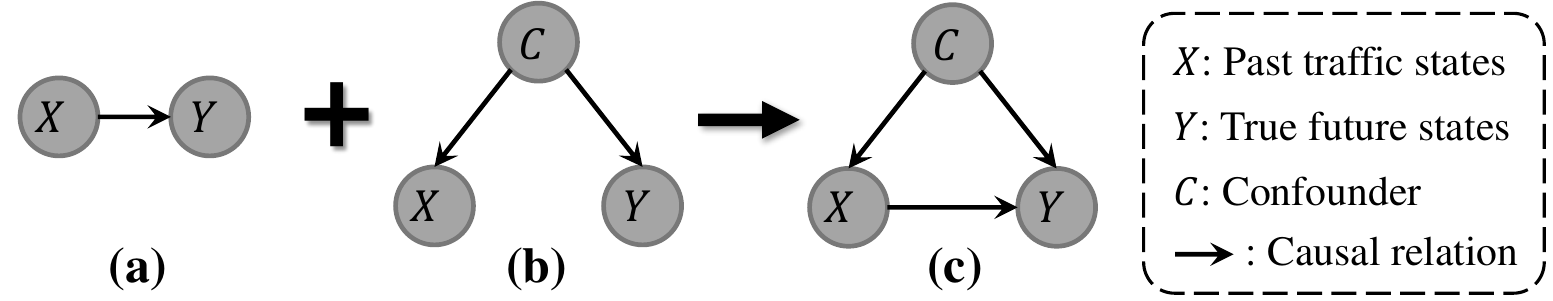}\vspace{-3mm}
    \caption{Structural causal model for traffic forecasting.}
    \label{fig:scm}\vspace{-5mm}
\end{figure}

While significant efforts have been made in previous works, most can be classified under the same modeling paradigm from the perspective of causal modeling, namely, modeling the directed causal relation $X \rightarrow Y$ (see \fig~\ref{fig:scm}(a)). This paradigm assumes a stable and direct causal relationship between $X$ and $Y$, allowing for effective modeling of this relationship through a data-driven approach. However, this assumption does not always hold in urban traffic systems. Spatiotemporal dependencies between $X$ and $Y$ can be influenced by various external factors such as rain, traffic accidents, holidays, and other events. In the causal modeling theory, these external factors can be expressed as confounders $C$, which simultaneously affect the states of $X$ and $Y$, causing shifts in the $X \rightarrow Y$ relationship (see \fig~\ref{fig:scm}(b)). This issue limits the generalization of existing traffic prediction models under extreme weather or emergency situations, compromising the resilience of cities. For instance, heavy snow (a type of confounder) can lead to more cautious driving behavior, resulting in severe congestion during non-peak hours and altering the relationship between $X$ and $Y$. If this changing relationship is ignored, the model cannot be expected to perform well in traffic management during snowy days.

Classic approaches in causal modeling theory to address the confounder issue include front-door adjustment and back-door adjustment~\cite{scm}. The front-door adjustment aims to identify a mediator variable that lies on the causal pathway between $X$ and $Y$ and is not influenced by any other confounders $C$. In contrast, the back-door adjustment controls the confounder $C$ to estimate the causal effect of $X \rightarrow Y$ under different confounder values. In recent years, these adjustment approaches have also been incorporated into deep learning models for traffic forecasting. These methods explore potential confounders or mediators and use deep learning to extract their representations to achieve deep learning-based front-door and back-door adjustments. For example, STNSCM~\cite{stnscm} uses time and location as mediators, learning their representations to achieve front-door adjustment for deep learning-based bike flow prediction. CaST~\cite{xia2023deciphering}, on the other hand, learns the representation of an environment codebook to implement back-door adjustment, removing the influence of environment confounders. It also represents spatial context as a mediator of front-door adjustment, thus eliminating the impact of spatial location confounders.

Although these methods have been effective in addressing the confounder issue, there remain two significant challenges that need to be solved in real-world traffic prediction applications. \textbf{First}, existing methods require traversing all possible values of confounders or mediators, necessitating that their values must be discrete. However, in real-world scenarios, many confounders and mediators are with continuous value. An approximate method is to quantize them as a discrete value~\cite{scm}. However, setting the correct quantization step size is very difficult. Too small a quantization step results in insufficient data for each condition, while too large a quantization step fails to fully eliminate the effects of confounders.
\textbf{Second}, existing methods require confounders or mediators to be predefined. However, traffic prediction is a complex open scenario that is influenced by many unknown factors that cannot be predefined~\cite{wang2019understanding}, such as periodicity and rhythm in time, spatial location and land function, and even some uncertainties, such as major events, weather and \etc. Therefore, it is very difficult to represent all possible confounders using an explicit way.

To overcome the above challenges, we propose a Spatial-Temporal sElf-superVised confoundEr learning ({\model}) model that adopts a self-supervised method to learn representations of implicit confounders in traffic forecasting. Our model employs a basis vector approach to address the challenge of traversing all possible confounders in back-door adjustment. Specifically, we use neural networks to learn a set of basis vectors for confounders, termed the base confounder bank, rather than targeting specific confounders. Using the base confounder bank, we can represent any confounder, whether continuous or discrete, predefined or not, as a linear combination of these basis vectors. The combination weights are adaptively produced by performing cross-attention between the input sample and the base confounder bank. Next, to ensure that the base confounder bank has adequate expressive capacity to handle various types of confounders, we propose three self-supervised auxiliary tasks for its training. The tasks include spatial location classification, temporal index identification, and traffic load prediction for incorporating spatial, temporal, and semantic information about confounders, respectively. Finally, we adopt a confounder-irrelevant relation decoupling module to separate the confounder effects from the direct $X \rightarrow Y$ relations. It includes an adversarial disentanglement component for semantic separation and mutual information minimization loss for distribution separation. The confounder representations from the base confounder bank and the $X \rightarrow Y$ relation representations are then transformed into corresponding traffic state predictions, followed by a fusion module that combines these predictions to generate the final results.

Extensive experiments on four large-scale traffic datasets demonstrate the superiority of our \model in scenarios with data distribution shifts due to spatial and temporal confounders. A further study on the weather confounder highlights our model's adaptability and generalizability to unseen confounders. 
To our knowledge, this is the first work to extend the principle of back-door adjustment to handle continuous or unknown confounders in deep-learning-based traffic prediction.


\section{Notation and Problem Definition}\label{sec:pre}

\subsection{Notation}

We use a traffic graph to model the dynamic states of urban traffic.

\begin{mydef}[Traffic Graph]
    Given a set of traffic entities (\eg spatial regions, road segments), denoted as $\mathcal{V} = \{v_n | 1 \le n \le N\}$, we define a traffic graph as $\mathcal{G} = \left(\mathcal{V}, {\bm A}\right)$. Here, $\bm{A}\in \mathbb{R}^{N \times N}$ is a binary adjacent matrix for the graph, where $a_{m,n} = 1$ when there is an edge from the node $v_m$ to $v_n$.
\end{mydef}

Over the traffic graph, we define dynamic traffic states.

\begin{mydef}[Traffic State]
    Given a traffic entity $v_n$, assuming it has $F$ traffic state features, such as average speed, traffic inflow, and traffic outflow, we denote the traffic states of $v_n$ at the $t$-th time slice as $\bm{x}_{n,t}\in \mathbb{R}^F$. The traffic states for all the $N$ entities are denoted as a matrix of $\bm{X}_t \in \mathbb{R}^{N\times F}$. The historical traffic states during $t-T+1$ to $t$ are expressed as $\bm{\mathcal{X}}_t = \left(\bm X_{t-T+1}, \dots, \bm X_{t} \right)\in \mathbb{R}^{T \times N \times F}$, and $\bm{Y}_{t+1} = \bm{X}_{t+1}$ denotes the future traffic states.
\end{mydef}


\subsection{Problem Definition}

Given past spatiotemporal (ST) traffic states $\bm{\mathcal{X}}_t$ and future states $\bm{Y}_{t+1}$, the classic traffic prediction problem is to find a function of
\begin{equation}\label{eq:problem-old}\small
  \hat{\bm{Y}}_{t+1} = f\left(\bm{\mathcal{X}}_t; \bm{\Theta}\right),
\end{equation}
where $f(\cdot)$ is the forecasting function, $\hat{\bm{Y}}_{t+1}$ is the prediction for $\bm{Y}_{t+1}$ and $\bm{\Theta}$ is the parameters to learn.

To reveal the underlying mechanism of traffic data dynamics, we adopt the Structural Causal Model (SCM)~\cite{scm} to describe the relations between the elements in the traffic prediction problem. 
We denote the history traffic states as $X$ and the future traffic states to be predicted as $Y$. 
There are two types of effects that can cause correlations between $X$ and $Y$ (as illustrated in \fig~\ref{fig:scm}):
\begin{enumerate}[leftmargin=*]
  \item The direct casual affection from $X$ to $Y$, denoted as $X \rightarrow Y$.
  \item A confounder $C$ which can effect both $X$ and $Y$ at the same time, \ie $X \leftarrow  C \rightarrow Y$.
\end{enumerate}

However, most existing works only model the first relation that is irrelevant to confounders, which can be denoted as $\mathrm{Pr}^{(i)}(Y|X)$. 
This is due to the problem definition in \equ~\eqref{eq:problem-old} does not explicitly describe the influence of the confounder $C$.
Such a definition induces researchers to ignore the effect of confounders in model design and limits the generalizability of the learned model.
If we consider such effects, the corresponding conditional probability $\mathrm{Pr}^{(c)}(Y|X)$ can be expressed as
\begin{equation}\label{eq:pr_v}\small
    \mathrm{Pr}^{(c)}(Y|X) = \mathrm{Pr}(Y|X,C) \mathrm{Pr}(C|X).
\end{equation}
Then, the traffic prediction problem relevant to confounders is divided into two steps. In the first step, we approximate $\mathrm{Pr}(C|X)$ via a learning model that aims to extract confounder representations from historical traffic state data:
\begin{equation}\label{eq:x2c}\small
    \bm{C}_t = g^{(c)}(\bm{\mathcal{X}}_t),
\end{equation}
where $g^{(c)}(\cdot)$ is the confounder representations extracting function. 
The second step employs the confounder representations and input data to predict future traffic states: 
\begin{equation}\small
  \hat{\bm{Y}}^{(c)}_{t+1} = f^{(c)}\left(\bm{\mathcal{X}}_t, \bm{C}_t, \bm{\Theta}^{(c)}\right),
\end{equation}
where $f^{(c)}(\cdot)$ is the confounder-related forecasting function and $\bm{\Theta}^{(c)}$ is its parameters.

By combining relations $\mathrm{Pr}^{(c)}(Y|X)$ and $\mathrm{Pr}^{(i)}(Y|X)$, we can derive the overall probability relation as $\mathrm{Pr}(Y|X) = \mathrm{Pr}^{(c)}(Y|X) + \mathrm{Pr}^{(i)}(Y|X)$. 
Therefore, the ST traffic prediction problem should be \textbf{reformulated} as
\begin{equation}\label{eq:problem}\small
    \hat{\bm{Y}}_{t+1} = f^{(c)}\left(\bm{\mathcal{X}}_t, \bm{C}_t; \bm{\Theta}^{(c)}\right)+ f^{(i)}\left(\bm{\mathcal{X}}_t; \bm{\Theta}^{(i)}\right),
\end{equation}
where $f^{(i)}(\cdot)$ is the confounder-irrelevant forecasting function with learnable parameters $\bm{\Theta}^{(i)}$.

\section{Model}\label{sec:method}

\begin{figure}[t]
    \centering
    \includegraphics[width=\columnwidth]{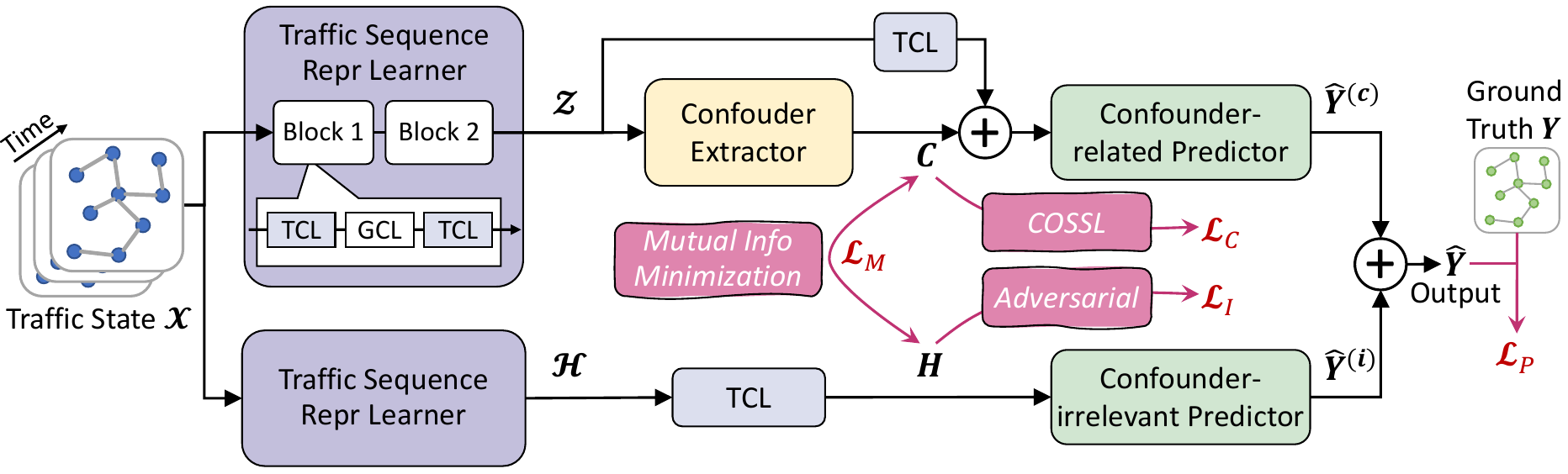}\vspace{-.3cm}
    \caption{The pipeline of our STEVE model. Repr: Representation. TCL: Temporal Convolutional Layer. GCL: Graph Convolutional Layer. Info: Information. COSSL: Confounder-Oriented Self-Supervised Learning. We omit the sample index of all variables for simplicity. 
    \fig~\ref{fig:ce} illustrates the details of the confounder extractor. 
    }\label{fig:framework}\vspace{-.4cm}
\end{figure}

We implement \equ~\eqref{eq:problem} by proposing \model depicted in \fig~\ref{fig:framework}.
Our model takes historical ST traffic observations as input to predict future traffic states, with the aid of self-supervised signals to facilitate confounder representation extraction. 
We will elaborate on the pipeline and each core component in the following parts. 

\subsection{Confounder Representation Generation}

The goal of this component is to generate the confounder representations through the input traffic data. To achieve this, we first utilize a \textit{Traffic Sequence Representation Learner} module to embed dynamic temporal dependencies and variant spatial relations into a hidden representation. 
Then, we introduce a learnable \textit{Confounder Extractor} to extract complex dynamic confounder representations from the hidden representation adaptively.
Lastly, the confounder representations will be refined to represent the desired confounders by \textit{Confounder-Oriented Self-Supervised Learning} in Section \ref{sec:cossl}.

\subsubsection{Traffic Sequence Representation Learner}\label{sec:tsrl}

The TSRL module aims to transform the input traffic sequence $\bm{\mathcal{X}}_t \in \mathbb{R}^{T\times N\times F}$ into a hidden representation $\bm{\mathcal{Z}}_t \in \mathbb{R}^{T\times N\times D}$. 
Temporal and graph convolutional layers are employed in TSRL to model temporal patterns and spatial dependencies between different locations. 

\paratitle{Temporal Convolutional Layer (TCL).} We take traffic state sequence $\bm{\mathcal{X}}_t = \left(\bm X_{t-T+1}, \dots, \bm X_{t} \right) \in \mathbb{R}^{T \times N \times F}$ as the input of the TCL. 
We employ 1D convolution~\cite{gwnet} along the time dimension to implement the TCL, which outputs time-aware traffic embeddings:
\begin{equation}\label{eq:tcl}\small
(\bm{E}_{t-T_1+1},\dots,\bm{E}_t) = \mathrm{TCL}(\bm{X}_{t-T+1},\dots,\bm{X}_t),
\end{equation}
where $\bm{E}_t \in \mathbb{R}^{N \times D}$ is the traffic embedding matrix at time step $t$, and $T_1$ is the length of the output sequence. Here, $N$ is the node number of our input network, and $D$ is the embedding dimension.


\paratitle{Graph Convolutional Layer (GCL).} We take the output of TCL as input. Our GCL is implemented by a graph-based message-passing network~\cite{ji2023spatio}:
\begin{equation}\small
    \bm{S}_{t} = \mathrm{GCL}(\bm{E}_t, \bm{A}),
\end{equation}
where $\bm{A}$ is the adjacency matrix of the corresponding network. 
By applying GCL to each time-aware representation $\bm{E}_{t}$, we obtain the refined traffic representations $(\bm{S}_{t-T_1+1}, \dots, \bm{S}_{t})$.

To jointly model temporal and spatial dependencies, we draw inspiration from \cite{stgcn} and construct TSRL by two blocks, each of which shows like: $\mathrm{TCL}\to\mathrm{GCL}\to\mathrm{TCL}$, as in \fig~\ref{fig:framework}.  
The final output of TSRL is $\bm{\mathcal{Z}}_t \in \mathbb{R}^{T\times N\times D}$:
\begin{equation}\small
\bm{\mathcal{Z}}_t = (\bm{Z}_{t-T+1}, \dots, \bm{Z}_{t}) = \mathrm{TSRL}(\bm{\mathcal{X}}_t, \bm{A}).
\end{equation}
We use $\bm{\mathcal{Z}}_t$ for the confounder representation extraction in the next part. 
Similarly, we also employ another TSRL model with different parameters to generate representation $\bm{\mathcal{H}}_t \in \mathbb{R}^{T \times N \times D}$, which is used for confounder-irrelevant relation modeling in Section~\ref{sec:con_irr}.

\subsubsection{Confounder Extractor}\label{sec:cfd_ex}

This section aims to implement function $g^{(c)}(\cdot)$ in \equ~\eqref{eq:x2c} that extracts
confounder representations from historical traffic state data.

\paratitle{Motivation and Idea.} The function $g^{(c)}(\cdot)$ is used for approximating $\mathrm{Pr}(C|X)$ in \equ~\eqref{eq:pr_v}, where $C$ and $X$ are random variables of confounder and historical traffic data.
However, directly approximating $\mathrm{Pr}(C|X)$ is a non-trivial task that involves two main obstacles:
(1) $C$ has a complex distribution mixing different conditions; 
(2) $C$ could take on an infinite number of values.
To address these challenges, we draw inspiration from \cite{reynolds2009gaussian} and introduce a series of base variables to represent it:
\begin{equation}\label{eq:c2x_beta}\small
    \mathrm{Pr}(C|X) = \sum_{k=1}^{K} \alpha^{(k)} \phi\left(\beta^{(k)}|X\right),
\end{equation}
where $K$ is the number of base variables, $\alpha_k$ is the membership degree that $C$ belongs to the $k$-th variable, and {\small $\sum_{k=1}^{K} \alpha^{(k)} = 1$}. 
{\small $\phi(\beta^{(k)}|X)$} denotes the probability function of the $k$-th base variable given $X$.
We can treat {\small $\phi(\beta^{(1)}|X), \dots, \phi(\beta^{(K)}|X)$} as different base conditions that form complex confounder with learnable weights {\small $\bm{\alpha} = (\alpha^{(1)}, \dots, \alpha^{(K)})$}.
Moreover, since the change in weights is continuous, we can theoretically express an infinite number of confounders. 
For example, suppose we have base conditions such as rush hours, rainfall, holidays, \etc, by assigning different weights to them, we can express any complex confounders such as ``rush hours on a rainy workday'', \etc.

\begin{figure}[t]
    \centering
    \includegraphics[width=0.9\columnwidth]{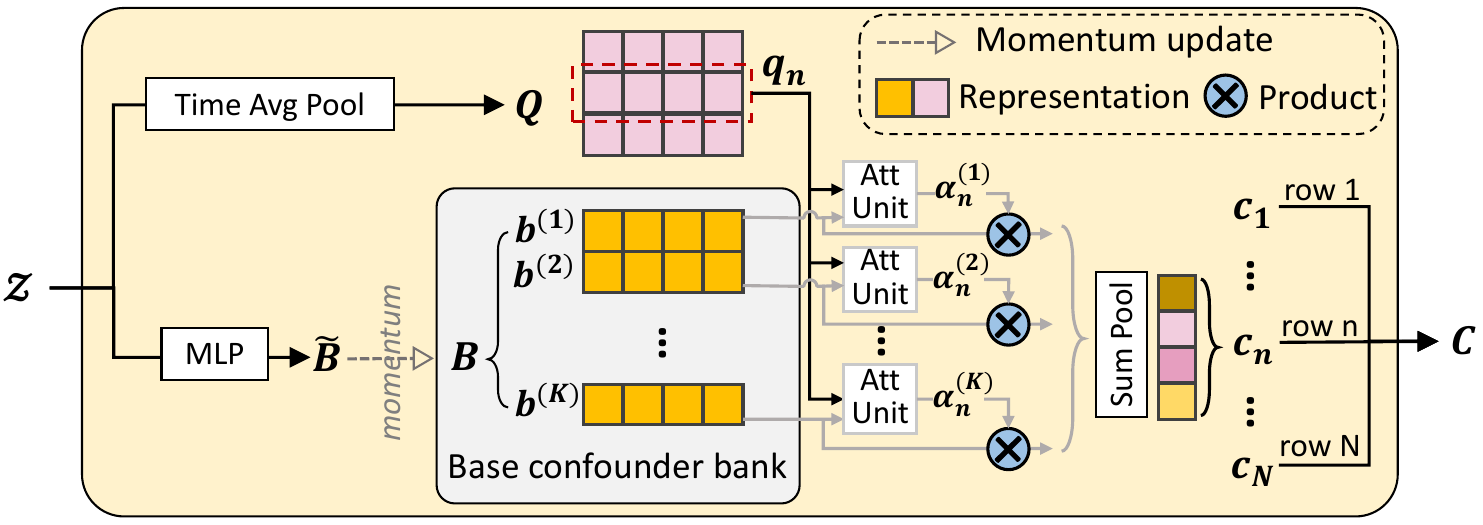}\vspace{-.3cm}
    \caption{The architecture of our confounder extractor. Avg: Average. Att: Attention. For simplicity, the sample index $t$ for $\bm{\mathcal{Z}}$ and $\bm{C}$ is omitted.}
    \label{fig:ce}\vspace{-.4cm}
\end{figure}

\paratitle{Method.} On top of this idea, as shown in \fig~\ref{fig:ce}, we implement an adaptive base confounder bank {\small $\bm{B} = (\bm{b}^{(1)}, \dots, \bm{b}^{(K)})^\top \in \mathbb{R}^{K \times D}$} with embedding dimension $D$. 
Here, {\small $\bm{b}^{(k)}$} is an implementation of {\small $\phi(\beta^{(k)}|X)$}. 
Specifically, after acquiring representation {\small $(\bm{\mathcal{Z}}_1, \dots, \bm{\mathcal{Z}}_t)$} in chronological order, we transform them into {\small $(\tilde{\bm{B}}_1, \dots, \tilde{\bm{B}}_t)$} by
\begin{equation}\label{eq:bank_pre}\small
    \tilde{\bm{B}}_t = \mathrm{MLP}(\mathrm{Flatten}_2(\bm{\mathcal{Z}}_t)).
\end{equation}
$\mathrm{Flatten}_2(\cdot)$ denotes the operation that flattens the first two dimensions of the input tensor $\bm{\mathcal{Z}}_t \in \mathbb{R}^{T\times N \times D}$.
$\mathrm{MLP}(\cdot)$ is employed in the first dimension of input data to generate $\tilde{\bm{B}}_t$ with shape $K \times D$.

Then, we use $\tilde{\bm{B}}$ to update $\bm{B}$.
A direct approach is real-time updating, \ie $\bm{B}_t = \tilde{\bm{B}}_t$, which preserves sufficient environmental information of the current traffic data sample but loses that of previous samples. This is inconsistent with our expectation that $\bm{B}_t$ should encompass more environmental information when perceiving the current environment.
To tackle this issue, we adopt a momentum update mechanism as follows:
\begin{equation}\small
    \bm{B}_t = \gamma \bm{B}_{t-1} + (1 - \gamma) \tilde{\bm{B}}_t,
\end{equation}
where $\gamma$ is the momentum coefficient to determine the amount of information being kept in every update.
Besides, we randomly initialized the confounder bank and utilized whitening with principal components analysis~\cite{abdi2010principal} to decorrelate the base confounder vectors.


Based on the confounder bank, we can generate the weight $\alpha^{(k)}$ in \equ~\eqref{eq:c2x_beta} via a cross attention mechanism\cite{vaswani2017attention}.
Concretely, for the $n$-th traffic entity $v_n$, we produce a confounder query $\bm{q}_{n,t} \in \mathbb{R}^{D}$ by 
\begin{equation}\small
    \bm{q}_{n,t} = \frac{1}{T} \sum_{\tau=t-T+1}^{t} \bm{z}_{n,\tau},
\end{equation}
where $\bm{z}_{n,\tau}$ is the hidden representation of $v_n$ at time slot $\tau$.
We then utilize the query $\bm{q}_{n,t}$ and the $k$-th vector in confounder bank $\bm{b}_t^{(k)}$ to compute $\alpha^{(k)}_{n,t}$ by:
\begin{equation}\small
    \alpha^{(k)}_{n,t} = \frac{\exp\left(u\big(\bm{q}_{n,t}, \bm{b}^{(k)}_t\big)\right)}{\sum_{j=1}^{K}\exp\left(u\big(\bm{q}_{n,t}, \bm{b}^{(j)}_t\big)\right)},
\end{equation}
where $u: \mathbb{R}^{D}\times \mathbb{R}^{D} \to \mathbb{R}$ is a function that computes attention weights, and we implement it by an MLP.
Lastly, we derive the relevant confounder embedding as 
\begin{equation}\label{eq:cfd_repr}\small
    \bm{c}_{n,t} = \sum_{k=1}^{K} \alpha_{n,t}^{(k)}\bm{b}_{t}^{(k)}.
\end{equation}
$\bm{c}_{n,t}$ is the $n$-th row of $\bm{C}_t \in \mathbb{R}^{N \times D}$, which is the confounder representation for input sample $\bm{\mathcal{X}}_t$ with hidden representation $\bm{\mathcal{Z}}_t$. 
When the $(t+1)$-th sample $\bm{\mathcal{X}}_{t+1}$ comes, we can similarly feed its $\bm{\mathcal{Z}}_{t+1}$ into \equ~\eqref{eq:bank_pre} and repeat the procedure until \equ~\eqref{eq:cfd_repr}.

\paratitle{Remark:} After the training phase, our base confounder bank $\bm{B}$ defines a vector space, where each confounder can be regarded as a point (in this space) that possesses a unique coordinate defined by weights $(\alpha^{(1)}, \dots, \alpha^{(K)})$.
The existence of an infinite number of points in space indicates that our base confounder bank can represent any possible confounder in the learned space, whether continuous or discrete, predefined or not. 
Furthermore, when testing, a new traffic sample can slightly shape the confounder space according to its hidden environment. 
This enhances the ability of our method to generalize to new unseen confounders.



\subsection{Confounder-Oriented SSL}\label{sec:cossl}

The Confounder-Oriented Self-Supervised Learning (COSSL) component aims to refine representation $\bm{C}$ by using self-supervised signals relevant to confounders. 
Since it is hard to enumerate all confounders explicitly, we propose to use some representative ones as self-supervised signals to inject confounder information into $\bm{C}$\footnote{For simplicity, we omit the sample index $t$ of $\bm{C}_t$.}. 

Specifically, we categorize potential factors that affect traffic states into three classes from conceptually different perspectives, \ie temporal, spatial, and semantic, based on the unique properties of ST traffic data. 
We then carefully select representative and easily collected confounders from each class, including temporal index, spatial location, and traffic capacity.
These selected factors will serve as self-supervised signals in the following three tasks. 

\textbf{Task \#1: Spatial Location Classification}. The spatial location of a traffic entity reflects its surroundings, which may appear as a confounder and vary by location, altering the dependency of past and future data (\eg $(\bm{x}_{t-T+1}, \dots, \bm{x}_{t}) \to \bm{x}_{t+1}$). 
For example, such dependency in a transportation hub can significantly differ from that in a working area. Therefore, we propose a spatial location classification task to perceive the surroundings of each region.
Firstly, for traffic entity $v_n \in \mathcal{V}$, we utilize the node ID to assign it a unique one-hot location label, $\bm{y}_{n}^{(1)} \in \{0, 1\}^{N}$, where its item ${y}_{n,m}^{(1)} = 1$ if $m = n$ else 0. 
We optimize the task by a cross-entropy loss as
\begin{equation}\label{eq:S}\small
    \ell_{sl}(\bm{C}) = \frac{1}{N}\sum_{n=1}^{N} \ell_{sl}^{(n)} = \frac{1}{N}\sum_{n=1}^{N} \sum_{m=1}^{N} {y}_{n,m}^{(1)}\log\left(\hat{{y}}_{n,m}^{(1)}\right),
\end{equation}
where $\hat{{y}}_{n,m}$ is the predicted probability of the $n$-th entity belonging to category $m$, and it is the $m$-th item of vector $\hat{\bm{y}}_{n}^{(1)} = g_{1}(\bm{c}_n) \in \mathbb{R}^{N}$.
$g_1(\cdot)$ is a two-layer MLP followed by a softmax activation, while $\bm{c}_n$ is the $n$-th row of confounder representation $\bm{C}$.

\textbf{Task \#2: Temporal Index Identification}. Time-varying confounders like weather and holidays can shape the traffic data distribution. 
For instance, holidays flatten the curves of morning and evening rush hours, resulting in a very different distribution from the workday rush hours.
To utilize such information, we propose a temporal index identification task.
Specifically, we divide the day into 24 time slots, each of which is a category. 
We use different categories to distinguish between workdays and holidays, so there is a total of $I_t=48$ temporal indexes. 
For a given traffic state sample $(\bm{\mathcal{X}}, \bm{Y})$, we use the temporal index of $\bm{Y}$ as ground truth. 
It is denoted by a one-hot vector $\bm{y}^{(2)}\in \{0, 1\}^{I_t}$. The optimization objective of the temporal index identification task is
\begin{equation}\label{eq:T}\small
    \ell_{ti}(\bm{C}) = \sum_{i=1}^{I_t}{y}_{i}^{(2)}\log\left(\sigma(\hat{\bm{y}}^{(2)})_i\right),
\end{equation}
where $\sigma$ is the SoftMax activation. 
{\small $\hat{\bm{y}}^{(2)} = \frac{1}{N} \sum_{n=1}^{N}g_{2}\left(\bm{c}_n\right)$} is the predicted temporal index vector, where $g_{2}$ is a two-layer MLP used for enhancing the confounder representation $\bm{c}_n$.


\textbf{Task \#3: Traffic Load Prediction}. The traffic load is a kind of semantic information describing the congestion level of traffic entities. 
It acts as a confounder and has an impact on the change of future traffic. 
For example, when the load reaches saturation, the traffic is more likely to be congested, causing traffic speeds and inflow/outflow to drop in subsequent time slots. 
Therefore, we propose a traffic load prediction task to inject dynamic load information into confounder representations.
Specifically, we approximate the load capacity of the $n$-th node by using the historical maximum traffic flow, \ie {\small $CP_n = \max(\{\bm{x}_{t,n}\}_1^{\tau})\in \mathbb{R}^{F}$}. 
$\tau$ denotes the number of time slots in the training set. 
$\max(\cdot)$ extracts the maximum value of each feature.
Then, we divide flow volume into 6 load levels and calculate the traffic load of the $n$-th node via {\small $\bm{y}_{n}^{(3)} = \lceil 5 {\bm{y}_ n}/{CP_n} \rceil \in \{0, \dots, 5\}^F$}, where $\bm{y}_ n$ is the label data in the main traffic prediction problem. 
Since load states are quite imbalanced in practice, we adopt the Mean Square Error (MSE) to optimize this task:
\begin{equation}\label{eq:L}\small
    \ell_{tl}(\bm{C}) = \frac{1}{N}\sum_{n=1}^{N} \left\Vert g_{3}\left(\bm{c}_{n}\right) - \bm{y}_{n}^{(3)} \right\Vert_2^2,
\end{equation}
where $g_{3}(\cdot)$ is the load prediction head implemented by a two-layer MLP, and $\bm{c}_{n}$ is the confounder representation.
It is worth noting that though quantized as a discrete value, traffic load has a relative size relationship. 
Regression loss like MSE is more suitable than classification loss since MSE can perceive size differences. 

Lastly, we jointly minimize all three self-supervised loss functions to train representation $\bm{C}$, making it fuse information of various confounders. 
The target loss of confounder-orient self-supervised learning is defined as
\begin{equation}\label{eq:c}\small
     \mathcal{L}_C = \sum_{u \in \{sl, ti, tl\}} \ell_u(\bm{C}).
\end{equation}
\noindent{\textbf{Remark:}} The selected representative factors that serve as self-supervised signals can instruct our model to effectively identify more information about latent confounders. 
By training auxiliary tasks that capitalize on these signals, we enhance our model to learn robust representations capable of previously unseen confounders.


\subsection{Confounder-Irrelevant Relation Decoupling}\label{sec:con_irr}

As introduced in \equ~\eqref{eq:problem}, in addition to capturing the confounder-related dynamic relationships, modeling confounder-irrelevant relationships is also crucial for spatiotemporal traffic prediction.
Since confounder-irrelevant relations should involve minimal information about the confounder, we propose to disentangle confounder-irrelevant representations and confounder representations from the semantics and distribution perspectives.






Recalling the hidden representation $\bm{\mathcal{H}}_t \in \mathbb{R}^{T \times N \times D}$ produced for confounder-irrelevant relation modeling in Section~\ref{sec:tsrl}, it is then transformed into $\bm{H}_t \in \mathbb{R}^{N \times D}$ by applying the TCL defined in \equ~\eqref{eq:tcl} along the temporal dimension $T$. 
Next, we elaborate on how to refine $\bm{H}$ into a confounder-irrelevant representation distinguished from the confounder representation $\bm{C}$. Note we omit the sample index $t$ of $\bm{H}_t$ and $\bm{C}_t$ for convenience.

\subsubsection{Adversarial Disentanglement}\label{sec:ad}

\begin{figure}
    \centering
    \includegraphics[width=0.8\columnwidth]{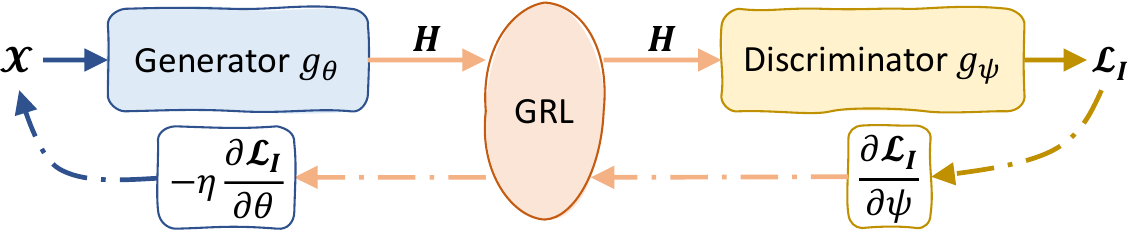}\vspace{-.3cm}
    \caption{Adversarial learning is achieved by inserting a GRL between the generator $g_\theta$ and the discriminator $g_\psi$. The forward pass is indicated by arrows while the backward pass is indicated by dashed arrows.}
    \label{fig:grl}\vspace{-.4cm}
\end{figure}

To push all confounder information away from $\bm{H}$, we introduce an adversarial learning-based disentanglement module as shown in \fig~\ref{fig:grl}.
Concretely, the generator $g_{\theta}$, consisting of a TSRL and a TCL, aims to produce $\bm{H}$.
It is then fed into the discriminator $g_{\psi}$ for confounder-related self-supervised tasks in Section~\ref{sec:cossl}.
Different from the learning pipeline of $\bm{C}$, we insert a Gradient Reversal Layer (GRL)~\cite{grl} between the generator and discriminator in the pipeline of $\bm{H}$ to disentangle $\bm{H}$ and $\bm{C}$.

The forward pass of GRL directly outputs the input without any transform: $\bm{H} = \mathrm{GRL}_{\eta}(\bm{H})$. 
However, during the backward pass, it multiples the incoming gradient back from the discriminator by a negative factor $-\eta$: 
\begin{equation}\small
    \frac{\partial~\mathrm{GRL}_\eta}{\partial~\bm{H}} = - \eta \bm{I},
\end{equation}
where $\bm{I}$ is the identity matrix.
The operation reverses the gradient direction passed back to the generator, pushing it away from the optimization direction of the confounder discriminator.
This results in $\bm{H}$ to be confounder-irrelevant, \ie the semantics of $\bm{H}$ in different confounder environments are as similar as possible. 

Mathematically, we can define the loss function that is being minimized as
\begin{equation}\label{eq:loss_i}\small
     \mathcal{L}_I(\psi, \theta) = \sum_{u \in \mathcal{U}} \ell_u(\mathrm{GRL}_{\eta}(\bm{H})) = \sum_{u \in \mathcal{U}} \ell_u(\mathrm{GRL}_{\eta}(g_{\theta}(\bm{\mathcal{X}}))),
\end{equation}
where $\mathcal{U} = \{sl, ti, tl\}$ is the task set of the self-supervised discriminator defined in Section~\ref{sec:cossl}, and $\psi$ is its parameters. 
Under the aforementioned loss function, the adversarial disentanglement module is actually trained with the following procedure:
\begin{equation}\small
    \psi = \mathop{\arg\min}_{\psi} ~\mathcal{L}_I(\psi, \theta),\quad
    \theta = \mathop{\arg\max}_{\theta} ~\mathcal{L}_I(\psi, \theta).
\end{equation}
That is, $\psi$ is optimized to minimize $\mathcal{L}_I$, and $\theta$ is optimized to maximize $\mathcal{L}_I$. 
With one single loss function, we achieve an adversarial relationship where two parts of the same network have different optimization objectives.
That's why we call it ``adversarial''.

\subsubsection{Mutual Information Minimization}\label{ssec:disent}

Generally, there is an independent constraint in the definition of disentangled representation~\cite{bengio2013representation}. 
To achieve this, we propose to minimize the Mutual Information (MI) between $\bm{H}$ and $\bm{C}$, making the representation distributions more disentangled:
\begin{equation}\label{eq:loss_m}\small
    \mathcal{L}_M = \mathrm{MI}(\bm{H}, \bm{C}) = \mathbb{E}_{\mathrm{Pr}(\bm{H}, \bm{C})}  \left[\log \frac{\mathrm{Pr}(\bm{H}, \bm{C})}{\mathrm{Pr}(\bm{H}) \mathrm{Pr}(\bm{C})}\right],
\end{equation}
where $\mathrm{Pr}(\bm{H}), \mathrm{Pr}(\bm{C})$, and $\mathrm{Pr}(\bm{H}, \bm{C})$ correspond to the marginal and joint distributions of $\bm{H}$ and $\bm{C}$.

Due to the unknown closed-form expressions of the marginal and joint distributions, direct computation of the MI in \equ~\eqref{eq:loss_m} is not feasible.
Therefore, we adopt an approximation method as an alternative.
Specifically, we use the CLUB method~\cite{club} to calculate the upper bound on MI of $\bm{H}$ and $\bm{C}$ as
\begin{equation}\label{eq:ub}\small
    \widehat{\mathrm{MI}}_{\mathrm{ub}}(\bm{H}, \bm{C}) = \frac{1}{M} \sum_{i=1}^{M} \bigg[\log q_{\theta}\left(\bm{H}_{i}|\bm{C}_i\right)
    -\frac{1}{M} \sum_{j=1}^{M}\log q_\theta\left(\bm{H}_{j}|\bm{C}_i\right)\bigg],
\end{equation}
where $M$ is the sample size, and $\bm{H}_{i}, \bm{C}_{i}$ denote representations produced by the $i$-th data sample. 
The {\small $q_{\theta}(\bm{H}|\bm{C})$} in \equ~\eqref{eq:ub} is a variational estimation of the conditional probability {\small $\mathrm{Pr}(\bm{H}|\bm{C})$}, which follows a Gaussian distribution as {\small $\mathcal{N}(\mu_{\bm{C}}|\sigma_{\bm{C}}^2)$}.
Here the mean $\mu_{\bm{C}}$ and variance $\sigma_{\bm{C}}^2$ are estimated using an MLP network:
\begin{equation}\small
    \{\mu_{\bm{C}},\sigma_{\bm{C}}^2\} = \mathrm{MLP}(\bm{C}, \bm{\Theta}_{\mathrm{mlp}}),
\end{equation}
where $\bm{\Theta}_{\mathrm{mlp}}$ refers the learnable parameters.

Finally, we use $\widehat{\mathrm{MI}}_{\mathrm{ub}}(\bm{H}, \bm{C})$ in \equ~\eqref{eq:ub} to replace $\mathrm{MI}(\bm{H}, \bm{C})$ in \equ~\eqref{eq:loss_m} to implement the mutual information minimization loss. 
This ensures that $\bm{H}$ and $\bm{C}$ adhere closely to the independent constraint in disentangled representation, which is empirically verified in Section \ref{appx:emb_vis}.

\subsection{Model Training}

In this section, we first make confounder-related and confounder-irrelevant predictions to compute the loss of the main traffic prediction task, and then combine it with other losses for model training. 

\paratitle{Prediction and Fusion.}
On the one hand, having $\bm{C}_t$ and $\bm{\mathcal{Z}}_t$, we can implement the confounder-aware forecasting function $f_c(\cdot)$ in \equ~\eqref{eq:problem} 
to make confounder-aware traffic predictions via 
\begin{equation}\label{eq:yc}\small
    \hat{\bm{Y}}^{(c)}_{t+1} = \mathrm{MLP}(\bm{C}_t + \mathrm{TCL}(\bm{\mathcal{Z}_t})).
\end{equation}
TCL, defined in \equ~\eqref{eq:tcl}, is responsible for absorbing the time dimension of $\bm{\mathcal{Z}}_t$. MLP is implemented by a two-layer fully connected network.
On the other hand, based on the confounder-irrelevant representation for the $t$-th sample, $\bm{H}_t$, we can implement the confounder-irrelevant forecasting function $f_i(\cdot)$ in \equ~\eqref{eq:problem} by 
\begin{equation}\small
    \hat{\bm{Y}}^{(i)}_{t+1} = \mathrm{MLP}(\bm{H}_t).
\end{equation}

In real scenarios, confounder factors affect regions to different degrees. 
Taking the rush hours factor as an example, it mainly affects the traffic states in office and residential areas, but exhibits less influence in parks and entertainment areas. 
Moreover, despite being in the same area, the influence on different state channels (\eg inflow, outflow) is also distinct.
Inspired by these phenomena, we propose a heterogeneity-aware fusion method as follows:
\begin{equation}\small
    \hat{\bm{Y}}_{t+1} = \bm{\Lambda_1} \odot \hat{\bm{Y}}^{(c)}_{t+1} + \hat{\bm{Y}}^{(i)}_{t+1},
\end{equation}
where $\hat{\bm{Y}}_{t+1}$ is the final traffic prediction. $\odot$ is the element-wise Hadamard product. $\bm{\Lambda_1} = \mathrm{Sigmoid}(\bm{C}_t \bm{W}_c)$ adjusts the confounder-related effect, where $\bm{W}_c \in \mathbb{R}^{D \times F}$ are learnable parameters.

\paratitle{Training Objective.}
Based on the final prediction $\hat{\bm{Y}}_{t+1}$, we can compute the main loss of the traffic prediction task by
\begin{equation}\label{eq:loss_p}\small
\mathcal{L}_P = \frac{1}{NF} \sum_{i=1}^{N}\sum_{j=1}^{F} \left\vert y_{i,j} - \hat{y}_{i, j} \right\vert,
\end{equation}
where $\hat{y}_{i, j}$ is the element of $\hat{\bm{Y}}_{t+1} \in \mathbb{R}^{N \times F}$, and $y_{i, j}$ denotes the ground truth. $N$ is the number of traffic entities, and $F$ is the number of traffic states being predicted. 
Finally, we obtain the overall loss by incorporating $\mathcal{L}_C$, $\mathcal{L}_M$, $\mathcal{L}_I$, and $\mathcal{L}_P$ into a joint learning objective:
\begin{equation}\small
    \mathcal{L}_O = \mathcal{L}_P + \gamma_1 \mathcal{L}_C + \gamma_2 \mathcal{L}_M + \gamma_3 \mathcal{L}_I,
\end{equation}
where $\gamma_1, \gamma_2, \gamma_3$ are the hyper-parameters to balance the learning of multiple tasks.

\section{Experiment}\label{sec:exp}

\begin{table*}[t]
  \centering
  \caption{Performance comparison of average on 5-run results. The bold/underlined font means the best/the second-best result. Work: Workday. Holi: Holiday. $c_i$: Spatial entity cluster with id $i$. Avg: Average results of different tasks.}\vspace{-.2cm}
  \resizebox{0.95\linewidth}{!}{
  \setlength{\tabcolsep}{0.4mm}
        \begin{tabular}{c|c|cc|c|cccc|c|cc|c|cccc|c|cc|cc}
    \toprule
    \multirow{3}{*}{Model} & Dataset & \multicolumn{8}{c|}{\textbf{NYCTaxi}}                                  & \multicolumn{8}{c|}{\textbf{NYCBike1}}                                 & \multicolumn{2}{c|}{\textbf{NYCBike2}} & \multicolumn{2}{c}{\textbf{BJTaxi}} \\
\cline{2-22}          &   \multirow{2}{*}{Task}    & \multicolumn{3}{c|}{TDS} & \multicolumn{5}{c|}{SDS} & \multicolumn{3}{c|}{TDS} & \multicolumn{5}{c|}{SDS} & TDS   & SDS   & TDS   & SDS \\
\cline{3-22}          &       & Work & Holi & Avg   & $c_0$    & $c_1$    & $c_2$ & $c_3$    & Avg   & Work & Holi & Avg   & $c_0$    & $c_1$    & $c_2$    & $c_3$    & Avg   & Avg   & Avg   & Avg   & Avg \\
        \midrule
    \multirow{2}[1]{*}{STGCN} & MAE   & 11.38  & 11.32  & 11.35  & 3.97  & 8.31  & 17.17  & 27.56  & 14.25  & 5.50  & 5.16  & 5.33  & 2.96  & 4.36  & 5.81  & {7.53} & {5.16 } & 5.48  & 5.30  & 12.14  & 15.80  \\
          & MAPE  & 18.90  & 18.69  & 18.80  & 27.68  & 16.80  & 11.42  & 9.74  & 16.41  & 25.28  & 29.98  & 27.63  & 33.86  & 28.50  & 24.56  & 22.28  & 27.30  & 27.90  & 28.04  & 17.13  & 14.31  \\
    \multirow{2}[0]{*}{AGCRN} & MAE   & \underline{10.87} & \underline{10.91} & \underline{10.89} & \underline{3.77} & \underline{8.17} & 17.12  & 27.68  & 14.19  & \underline{5.44} & \underline{5.06} & \underline{5.25} & 2.93  & \underline{4.35} & 5.87  & 7.56  & 5.18  & \underline{5.39} & \underline{5.18} & \underline{11.55} & 18.15  \\
          & MAPE  & \underline{18.28} & \underline{17.99} & \underline{18.14} & 26.14  & \underline{16.76} & \underline{11.34} & 9.65  & \underline{15.97} & \underline{25.19} & 29.71  & 27.45  & 33.46  & 28.88  & 25.41  & 22.61  & 27.59  & 27.39  & \underline{27.51} & 16.83  & 15.55  \\
    \multirow{2}[0]{*}{ASTGNN} & MAE   & 10.99  & 11.28  & 11.13  & 4.35  & 8.50  & 17.64  & 27.87  & 14.59  & 5.69  & 5.31  & 5.50  & 3.40  & 4.72  & 6.38  & 8.21  & 5.68  & 5.43  & 5.29  & 11.56  & \underline{15.07} \\
          & MAPE  & 19.45  & 24.27  & 21.86  & 33.39  & 17.44  & 11.48  & \underline{9.63} & 17.98  & 25.34  & 28.82  & 27.08  & 34.18  & 28.55  & 25.19  & 22.93  & 27.71  & 31.70  & 29.18  & 17.54  & \underline{14.08} \\
    \multirow{2}[0]{*}{HimNet} & MAE   & 13.42  & 13.16  & 13.29  & 4.36  & 9.59  & 20.25  & 33.64  & 16.96  & 5.98  & 5.48  & 5.73  & 3.18  & 4.63  & 6.15  & 8.07  & 5.51  & 5.49  & 5.27  & 12.04  & 16.08  \\
          & MAPE  & 20.78  & 20.23  & 20.50  & 29.18  & 18.80  & 13.16  & 10.98  & 18.03  & 26.28  & 29.97  & 28.13  & 31.98  & 28.84  & 26.00  & 23.88  & 27.68  & 27.96  & 27.66  & \underline{16.72} & 14.11  \\
          \midrule
    \multirow{2}[0]{*}{COST} & MAE   & 13.14  & 13.02  & 13.08  & 4.70  & 15.10  & 38.92  & 67.02  & 31.44  & 6.64  & 6.34  & 6.49  & 3.13  & 4.98  & 7.28  & 9.24  & 6.16  & 7.28  & 6.72  & 13.96  & 18.68  \\
          & MAPE  & 32.80  & 30.37  & 31.59  & 31.11  & 33.55  & 34.19  & 33.07  & 32.98  & 29.67  & 37.62  & 33.65  & 31.43  & 32.16  & 33.65  & 29.67  & 31.73  & 35.28  & 34.05  & 19.76  & 17.15  \\
    \multirow{2}[0]{*}{ST-Norm} & MAE   & 16.57  & 17.13  & 16.85  & 6.01  & 13.09  & 23.92  & 39.05  & 20.51  & 5.46  & 5.48  & 5.47  & 3.40  & 4.42  & \underline{5.76} & 7.98  & 5.39  & 5.48  & 5.17  & 13.31  & 17.05  \\
          & MAPE  & 31.47  & 30.55  & 31.01  & 45.77  & 30.53  & 18.23  & 16.37  & 27.72  & 25.46  & \underline{26.45} & \underline{25.96} & 33.29  & 27.42  & 24.31  & 21.26  & 26.57  & \underline{26.94} & 27.87  & 17.86  & 15.13  \\
    \multirow{2}[0]{*}{STWA} & MAE   & 14.13  & 14.58  & 14.36  & 4.09  & 9.97  & 23.45  & 37.35  & 18.72  & 6.90  & 6.54  & 6.72  & 3.73  & 5.37  & 7.27  & 9.46  & 6.46  & 9.44  & 9.08  & 13.25  & 17.55  \\
          & MAPE  & 21.06  & 21.08  & 21.07  & 28.11  & 19.79  & 15.24  & 13.09  & 19.05  & 28.70  & 32.11  & 30.41  & 37.60  & 31.66  & 27.70  & 25.65  & 30.65  & 44.11  & 40.83  & 18.69  & 15.66  \\
    \multirow{2}[1]{*}{SCNN} & MAE   & 12.62  & 12.78  & 12.70  & 4.19  & 9.26  & 20.27  & 31.58  & 16.33  & 6.55  & 6.16  & 6.36  & 3.48  & 5.13  & 7.03  & 8.76  & 6.10  & 5.71  & 5.54  & 12.24  & 16.13  \\
          & MAPE  & 21.29  & 20.88  & 21.09  & 29.29  & 19.45  & 14.49  & 12.17  & 18.85  & 27.49  & 32.90  & 30.20  & 34.73  & 31.10  & 27.90  & 24.84  & 29.64  & 28.62  & 28.44  & 17.31  & 14.65  \\
    \midrule
    \multirow{2}[1]{*}{AdaRNN} & MAE   & 15.16  & 16.96  & 16.06  & 4.81  & 17.97  & 29.20  & 35.25  & 21.81  & 7.22  & 6.13  & 6.68  & 3.27  & 5.45  & 8.00  & 10.35  & 6.77  & 5.96  & 8.71  & 18.71  & 25.77  \\
          & MAPE  & 41.49  & 32.21  & 36.85  & 35.52  & 35.59  & 40.89  & 46.73  & 39.68  & 29.64  & 33.34  & 31.49  & 30.80  & 32.83  & 31.57  & 28.52  & 30.93  & 32.51  & 39.62  & 25.34  & 22.32  \\
    \multirow{2}[0]{*}{CIGA} & MAE   & 15.23  & 15.34  & 15.29  & 5.32  & 8.76  & 19.19  & 27.65  & 15.23  & 6.47  & 5.76  & 6.12  & 3.21  & 4.73  & 6.72  & 9.07  & 5.93  & 5.96  & 6.17  & 13.08  & 17.85  \\
          & MAPE  & 18.95  & 21.51  & 20.23  & 27.11  & 17.02  & 13.33  & 16.09  & 18.39  & 29.27  & 32.61  & 30.94  & 34.82  & 28.69  & 25.08  & 21.99  & 27.64  & 29.97  & 31.97  & 19.48  & 16.70  \\
    \multirow{2}[0]{*}{STNSCM} & MAE   & 14.69  & 14.95  & 14.82  & 4.75  & 8.54  & 17.72  & \underline{25.69} & \underline{14.18} & 5.97  & 5.29  & 5.63  & 3.03  & 4.60  & 5.86  & 8.37  & 5.47  & 5.96  & 6.54  & 12.55  & 16.59  \\
          & MAPE  & 23.63  & 23.39  & 23.51  & \underline{24.11} & 22.65  & 12.43  & 11.11  & 17.58  & 26.67  & 29.91  & 28.29  & 26.84  & \underline{27.00} & \underline{23.28} & \underline{20.24} & \underline{24.34} & 29.51  & 28.33  & 18.16  & 15.67  \\
    \multirow{2}[0]{*}{CauSTG} & MAE   & 16.08  & 15.61  & 15.85  & 6.95  & 15.85  & 36.19  & 65.90  & 31.22  & 8.01  & 5.67  & 6.84  & 4.07  & 6.00  & 7.97  & 9.61  & 6.91  & 6.38  & 7.18  & 21.62  & 27.90  \\
          & MAPE  & 31.98  & 30.22  & 31.10  & 42.07  & 29.67  & 22.23  & 19.71  & 28.42  & 29.86  & 29.53  & 29.70  & 38.46  & 32.83  & 27.33  & 24.12  & 30.69  & 28.80  & 31.61  & 24.71  & 21.18  \\
    \multirow{2}[1]{*}{CaST} & MAE   & 13.15  & 14.36  & 13.76  & 4.89  & 9.87  & \underline{16.90} & 26.91  & 14.64  & 5.70  & 5.99  & 5.84  & \underline{2.71} & 4.75  & 6.18  & 8.15  & 5.45  & 6.72  & 6.43  & 12.35  & 16.44  \\
          & MAPE  & 19.53  & 18.44  & 18.99  & 31.84  & 20.30  & 14.55  & 11.84  & 19.63  & 26.31  & 29.43  & 27.87  & \underline{26.80} & 29.77  & 26.91  & 24.69  & 27.04  & 32.40  & 32.32  & 18.87  & 15.50  \\
    \midrule
    \multirow{2}[2]{*}{STEVE} & MAE   & \textbf{10.43} & \textbf{10.51} & \textbf{10.47} & \textbf{3.47} & \textbf{8.02} & \textbf{16.52} & \textbf{25.61} & \textbf{13.40} & \textbf{5.13} & \textbf{4.73} & \textbf{4.93} & \textbf{2.21} & \textbf{4.13} & \textbf{5.59} & \textbf{7.09} & \textbf{4.75} & \textbf{4.87} & \textbf{4.64} & \textbf{10.94} & \textbf{14.47} \\
          & MAPE  & \textbf{16.46} & \textbf{16.06} & \textbf{16.26} & \textbf{21.93} & \textbf{16.20} & \textbf{11.08} & \textbf{9.36} & \textbf{14.64} & \textbf{22.89} & \textbf{26.13} & \textbf{24.51} & \textbf{24.44} & \textbf{26.89} & \textbf{23.17} & \textbf{20.06} & \textbf{23.64} & \textbf{22.94} & \textbf{22.40} & \textbf{16.46} & \textbf{13.61} \\
    \bottomrule
    \end{tabular}%
    }\vspace{-.2cm}
  \label{tab:main}%
\end{table*}%

\subsection{Experimental Setting} 

\subsubsection{Dataset and Baseline} To evaluate our proposed method, we conduct experiments on four real-world traffic datasets including NYCTaxi, NYCBike1, NYCBike2, and BJTaxi~\cite{ji2023spatio}, which record the bike rental demands and taxi orders, respectively. We divide all datasets into training, validation, and test sets in a ratio of 7:1:2.

We choose Mean Absolute Error (MAE) and Mean Absolute Percentage Error (MAPE) as evaluation metrics, which are widely used in ST traffic prediction~\cite{stgcn,agcrn,deng2023spatio}.
A lower metric value indicates a better performance. 
We selected 13 methods as baselines and categorized them into distinct groups: $i)$ \underline{Spatiotemporal} \underline{prediction methods based on GNNs:} STGCN\cite{stgcn}, GMAN\cite{gman}, AST- GNN\cite{astgnn}, and HimNet\cite{himnet}; $ii)$ \underline{Disentanglement-based spatiotem-} \underline{poral methods:} COST\cite{woo2022cost}, ST-Norm\cite{deng2021st}, STWA\cite{fang2023stwave}, and SCNN\cite{deng2024scnn}; 
$iii)$ \underline{Models considering distribution shift:} AdaRNN\cite{adarnn},  CIGA\cite{chen2022learning}, STNSCM\cite{deng2023spatio}, CauSTG\cite{zhou2023maintain}, and CaST\cite{xia2023deciphering}.
The final model parameters are chosen by the optimal effect of the validation set.
Detailed descriptions of datasets and baselines are in Appendix~\ref{appx:setting}.

\subsubsection{Implementation Protocols} Our \model is implemented with PyTorch 1.10.2 on an Ubuntu server with an NVIDIA RTX 3090. 
Both temporal and spatial convolution kernel sizes in TSRL are set to 3. 
The hidden dimension $D$ is searched over $\{16, 32, 64, 128\}$. 
For the base confounder number $K$, we search it from $\{16, 32, 64, 128, 256\}$. 
For the momentum coefficient $\gamma$ in the confounder extractor, we test it from 0.1 to 0.9.
Our model is trained using Adam optimizer with a learning rate of 0.001 and a batch size of 32.
Task balancing coefficients $\gamma_1, \gamma_2, \gamma_3$ are trained via a dynamic weight-averaging strategy~\cite{liu2019end} with initial values 1.0.
Detailed model setting and parameter sensitivity are in Appendix~\ref{appx:setting} and \ref{appx:para_sen}.
Since hidden confounder data are unavailable, we assess the model's robustness on distribution shift via simulated environments. Specifically, we consider two scenarios that are common in the real world: 
(1) \textbf{Temporal Distribution Shift (TDS)}: we split the temporal distribution into workdays and holidays, which is roughly 5:2 in the training set. It is then shifted to 1:0 and 0:1 to imitate TDS to the maximum extent. 
(2) \textbf{Spatial Distribution Shift (SDS)}: To simulate real-world semantics of traffic entities, we cluster them into different groups via $k$-means algorithm. $\{c_0, \dots, c_k\}$ denote the clustering results, where entities with smaller id are usually located in less popular areas and thus have lower traffic. There is a mixed distribution consisting of all clusters in the training set, and we process the distribution so that it contains only one cluster, thus realizing SDS.

\subsection{Overall Performance}

We run all models five times and report the mean results in \tab~\ref{tab:main}. 
The details of NYCBike2 and BJTaxi are in Appendix~\ref{appx:main_tab}.
From \tab~\ref{tab:main}, we have four key findings: 
\textbf{(1)} \model consistently outperforms all competing baselines across every task on four datasets (according to the Nemenyi test at level 0.05 in Appendix~\ref{appx:sig_test}), while the second-best model is not consistent across all cases. This shows that \model offers more stable and reliable results, highlighting its robustness and adaptability to various distribution-shift scenarios.
\textbf{(2)} There is no significant uplift of unsupervised disentanglement-based methods \wrt classical ST prediction methods, indicating that decoupling without supervised signals does not effectively improve the model performance. That is why we incorporate self-supervised signals with disentanglement. 
\textbf{(3)} Some of the models against distribution shift yield unsatisfactory results. For instance, AdaRNN and CIGA fail to fully capture spatial and temporal dependencies. Meanwhile, CauSTG primarily focuses on learning invariant relations across different confounders, overlooking the importance of capturing variant relations in spatiotemporal prediction. Additionally, CaST's forced discretization of a continuous temporal environment disrupts the intrinsic structure of spatiotemporal data, increasing modeling difficulty. This confirms our model's effectiveness in modeling dynamic confounders in a basis vector approach without the need for discretization.
\textbf{(4)} While baseline models such as AGCRN and STNSCM can achieve runner-up MAE performance in certain cases, they exhibit a large margin in MAPE compared to \model. This demonstrates that \model not only delivers small absolute error but also showcases superior relative error across different cases. 
The relative error is usually a better indicator of the model's generalizability to various cases than the absolute one as it allows assessment of the precision of a result independently of the data scale.
In addition, our model also achieves decent training efficiency and scalability (see Section \ref{appx:eff} and \ref{appx:scala}), making it well-suited for practical applications in real-world scenarios.

\begin{table}[t]
  \centering
  \caption{Ablation study of \model on average MAE.} \vspace{-3mm}
  \resizebox{0.85\columnwidth}{!}{
  \setlength{\tabcolsep}{0.4mm}
    \begin{tabular}{l|cc|cc|cc|cc}
    \toprule
    Dataset & \multicolumn{2}{c|}{NYCTaxi} & \multicolumn{2}{c|}{NYCBike1} & \multicolumn{2}{c|}{NYCBike2} & \multicolumn{2}{c}{BJTaxi} \\
    \midrule
    Metric & TDS   & SDS   & TDS   & SDS   & TDS   & SDS   & TDS   & SDS \\
    \midrule
    \textbf{STEVE} & \textbf{10.47} & \textbf{13.40} & \textbf{4.93} & \textbf{4.75} & \textbf{4.87} & \textbf{4.64} & \textbf{10.94} & \textbf{14.47} \\
    (a) w/o cfd & 10.65  & 13.67  & 5.04  & 4.85  & 5.04  & 4.79  & 11.13  & 14.81  \\
    (b) w/o ssl & 11.70  & 15.01  & 4.96  & 4.77  & 5.06  & 4.94  & 11.78  & 15.56  \\
    (c) w/o ad & 10.54  & 13.52  & 4.97  & 4.79  & 4.98  & 4.75  & 11.11  & 14.73  \\
    (d) w/o mi & 10.72  & 13.69  & 4.96  & 4.78  & 5.03  & 4.80  & 11.00  & 14.55  \\
    \bottomrule
    \end{tabular}%
    }\vspace{-.3cm}
  \label{tab:abl}
\end{table}%

\begin{figure}[t]
    \centering
    \includegraphics[width=0.95\columnwidth]{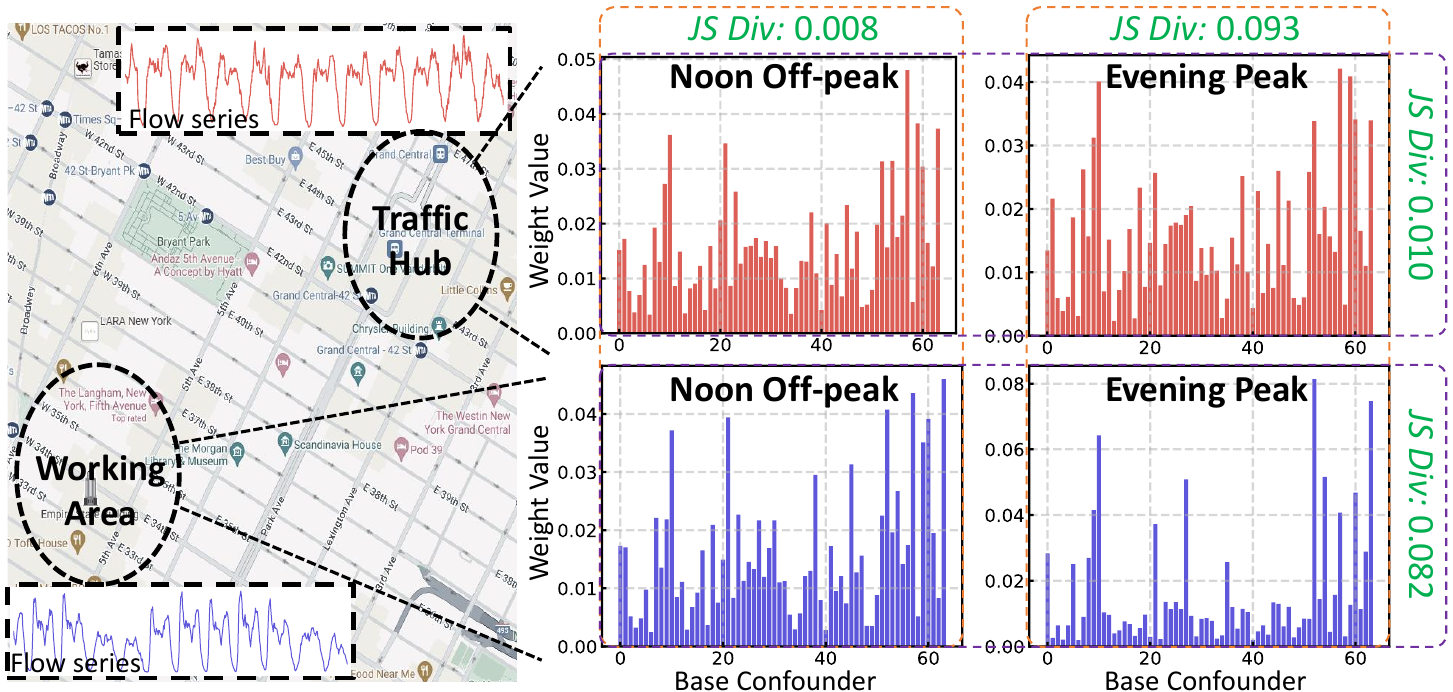}\vspace{-.3cm}
    \caption{Confounder distribution of distinct locations at different time periods. JS Div: Jensen–Shannon divergence.}\label{fig:cfd_dist}\vspace{-4mm}
\end{figure}

\subsection{Further Analysis of \model}

\subsubsection{Ablation Study}
To verify our model design, we carry out ablation experiments on the following variants:
(a) \textbf{w/o cfd} removes the confounder bank and takes $\bm{Q}$ as the confounder representation;
(b) \textbf{w/o ssl} removes the SSL tasks in \equ~\eqref{eq:c}; 
(c) \textbf{w/o ad} disables the adversarial disentanglement module in \equ~\eqref{eq:loss_i};
(d) \textbf{w/o mi} does not use the mutual information regularization in \equ~\eqref{eq:loss_m}. 
The MAE results of all datasets are shown in \tab~\ref{tab:abl}.
We can observe that all four components contribute to the model's overall performance.
Specifically, variants (a) and (b) show a great decrease, indicating that our proposed confounder bank can effectively extract confounder representation from limited observed ST data with the aid of SSL injecting representative confounder information. 
Besides, the impact of removing these components on performance is more pronounced for NYCTaxi and BJTaxi compared to NYCBike1 and NYCBike2 as taxi data possess more complex spatiotemporal relations and are more sensitive to confounder modeling.

\subsubsection{Analysis of Confounder Learning}
As introduced in Section \ref{sec:cfd_ex}, our confounder extractor can generate meaningful confounder embedding for each sample via a unique weight vector $\bm{\alpha}$ and confounder bank. To verify this, we visualize the weight distribution of different samples from NYCTaxi in \fig~\ref{fig:cfd_dist}.
We have two key observations: 
(1) The divergence in the weight distributions for the working area and traffic hub (especially in peak hours) suggests that our model learned the differences in the environments corresponding to these two types of zones.
(2) From noon off-peak to evening peak, the distribution of working areas gradually concentrates on a few base confounders, while that of traffic hubs remains dispersed. This shows that the learned environments at traffic hubs are consistently complex while working areas can be more regular in the evening peak, highlighting the ability of our model to capture the characteristics of both types of zones.

\subsubsection{Generalization to Unseen Confounders}
\begin{figure}[t]
    \centering
    \includegraphics[width=0.9\columnwidth]{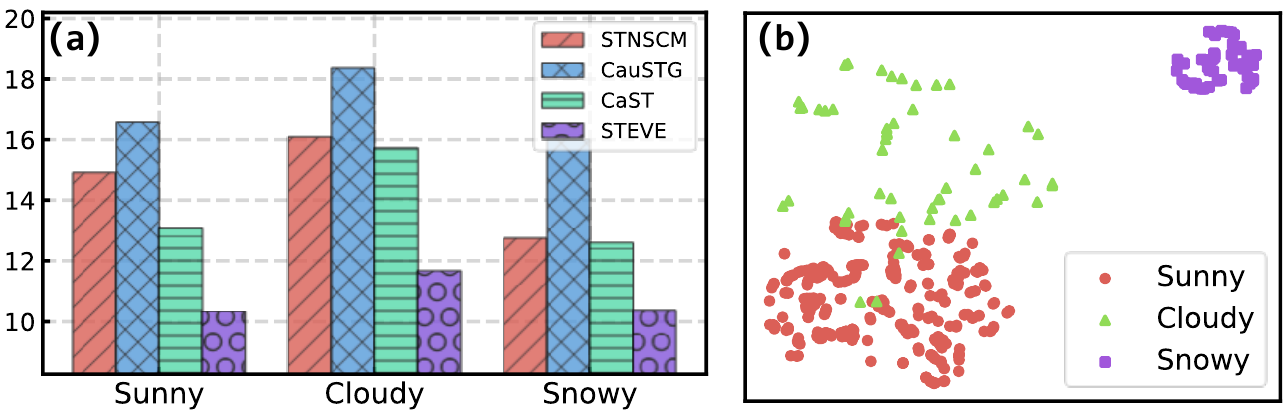}\vspace{-.3cm}
    \caption{(a) Unseen confounder generalization \wrt MAE. (b) The learned representations of corresponding confounders.}\label{fig:wea}\vspace{-5mm}
\end{figure}

In the COSSL module in Section~\ref{sec:cossl}, we subtly selected representative self-supervised signals to guide our model in capturing information about latent confounders, thereby improving the robustness and generalizability of the model in these latent environments.
To evaluate this, we collect the weather data of NYCTaxi, which is not exposed to model training.
We compare \model with three spatiotemporal models that do not employ self-supervised signals in solving the distribution shift problem, and the results are displayed in \fig~\ref{fig:wea}(a).
We can observe that \model consistently beats other baselines in all three weather conditions. 
This demonstrates the effectiveness of self-supervised signals' enhancement in confounder representations, making them robust to confounders not seen before.
Next, we dive into the weather confounder representations learned by our model.
The weights of these confounders are scattered in \fig~\ref{fig:wea}(b) by using t-SNE algorithm~\cite{van2008visualizing}.
From the results, we have two findings: 
(1) The confounder representations for the same type of weather are relatively close. Moreover, the representations of snowy days are more compact compared to those of sunny days, indicating snowy days characterize a more homogeneous confounder environment.
(2) In the representation space, the sunny confounder is closer to cloudy and farther from snowy. This highlights our model’s capability to extract informative confounder representations with the guidance of self-supervised signals.

\begin{figure}[t]
    \centering
    \subfigure[w/o MIM ($S=0.006$)]{\includegraphics[width=0.4\columnwidth]{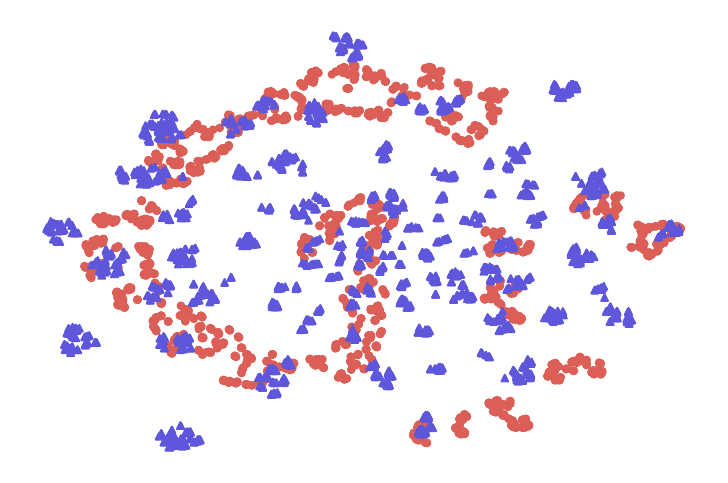}}\quad
    \subfigure[w/ MIM ($S=0.188$)]{\includegraphics[width=0.4\columnwidth]{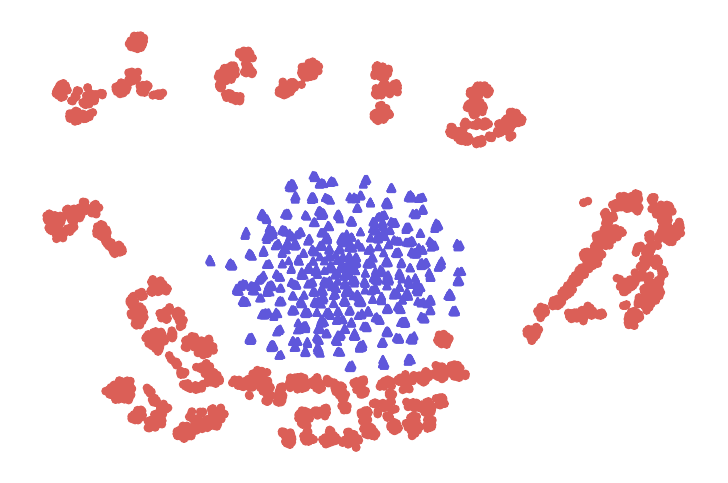}}\vspace{-3mm}
    \caption{Visualization of confounder-related representation $\bm{C}$ (red circle marker) and confounder-irrelevant representation $\bm{H}$ (purple triangle marker). $S$ is Silhouette Score.}\label{fig:womi}\vspace{-4mm}
\end{figure}

\subsubsection{Representation Visualization}\label{appx:emb_vis}

As introduced in Section~\ref{ssec:disent}, we propose to use Mutual Information Minimization (MIM) loss to disentangle representations $\bm{H}$ and $\bm{C}$ from the distribution perspective. 
To verify this, we randomly took some samples from the test set to generate the corresponding $\bm{H}$ and $\bm{C}$, and then used the t-SNE algorithm~\cite{van2008visualizing} to convert them as two-dimensional embedding vectors for visualization. 
As depicted in \fig~\ref{fig:womi}, \textbf{w/o MIM} and \textbf{w/ MIM} denote the results without and with MIM loss, respectively.
It can be seen that MIM facilitates the separation of distribution of $\bm{H}$ and $\bm{C}$ in the representation space, thus enhancing distributional independence for disentangled representations.

\begin{table}[t]
    \centering
    \caption{Computation time cost investigation by training/inference time per epoch in seconds.}
    \vspace{-3mm}
    \resizebox{0.85\columnwidth}{!}{
    \renewcommand{\arraystretch}{1}
    \begin{tabular}{c|cccc}
    \toprule
        Methods & NYCTaxi & NYCBike1 & NYCBike2 &  BJTaxi \\
        \midrule
        STGCN & 1.39/0.05 & 1.07/0.07 & 1.37/0.06 & 26.22/1.39 \\
        AGCRN & 25.94/6.34 & 9.82/1.24 & 25.23/5.95 & 68.22/13.05 \\
        ASTGNN & 12.91/1.14 & 9.48/0.74 & 10.14/0.96 & 35.16/3.43 \\
        HimNet & 5.43/1.27 & 9.26/1.96 & 5.63/1.28 & 53.37/9.63 \\
        \midrule
        COST & 3.21/1.42 & 3.41/1.82 & 3.27/1.37 & 6.32/5.98 \\ 
        ST-Norm & 5.28/0.37 & 9.62/0.48 & 5.02/0.36 & 12.98/0.78 \\ 
        STWA & 13.40/2.43 & 13.62/1.32 & 12.53/2.23 & 53.61/10.90 \\
        SCNN & 15.34/3.54 & 14.96/1.91 & 14.72/2.74 & 65.03/7.41 \\
        \midrule
        AdaRNN & 16.92/2.43 & 8.83/1.13 & 16.32/2.01 & 55.78/8.26 \\ 
        CIGA & 17.36/0.95 & 16.73/0.67 & 16.89/0.87 & 61.06/3.34 \\ 
        STNSCM & 1.66/0.19 & 2.79/0.32 & 1.74/0.19 & 4.94/0.68 \\ 
        CauSTG & 3.22/1.61 & 4.92/2.33 & 3.15/1.57 & 10.62/4.30 \\
        CaST & 25.31/7.20 & 37.82/8.13 & 21.07/6.52 & 81.93/18.73 \\
        \midrule
        STEVE & 4.51/0.52 & 3.83/0.45 & 2.22/0.31 & 27.13/2.98 \\
    \bottomrule
    \end{tabular}
    }
    \label{tab:eff_agcrn}\vspace{-4mm}
\end{table}

\subsubsection{Model Efficiency}\label{appx:eff}
In this section, we assess the efficiency of our model. Specifically, we measure the per-epoch training/inference time of all methods on all datasets, and the results are summarized in \tab~\ref{tab:eff_agcrn}.
To ensure fairness, all experiments are conducted on an Ubuntu server with an NVIDIA RTX 3090 with the same batch size.
From the results, we can observe that our model reduces the training and inference time by 73.7\% and 81.9\% on average compared to the best baseline AGCRN. While some baselines such as STGCN and COST surpass our model in time cost, our model achieves a win-win situation in terms of performance and training efficiency by combining the performance results in \tab~\ref{tab:main}.

\subsubsection{Model Scalability}\label{appx:scala}
\begin{figure}[tb]
    \centering
    \subfigure{
    \includegraphics[width=0.38\columnwidth]{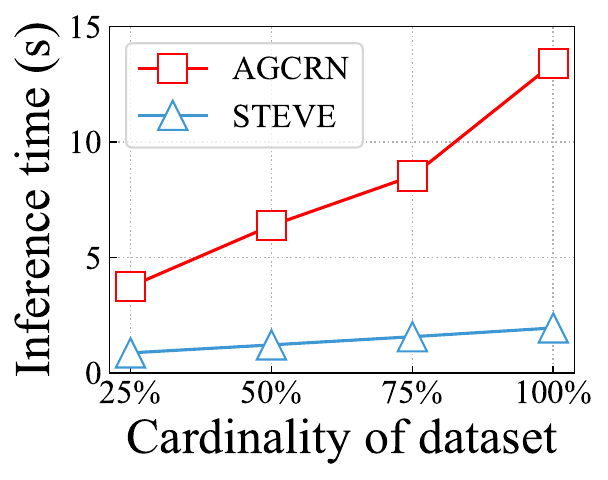}
    }\quad
    \subfigure{
    \includegraphics[width=0.38\columnwidth]{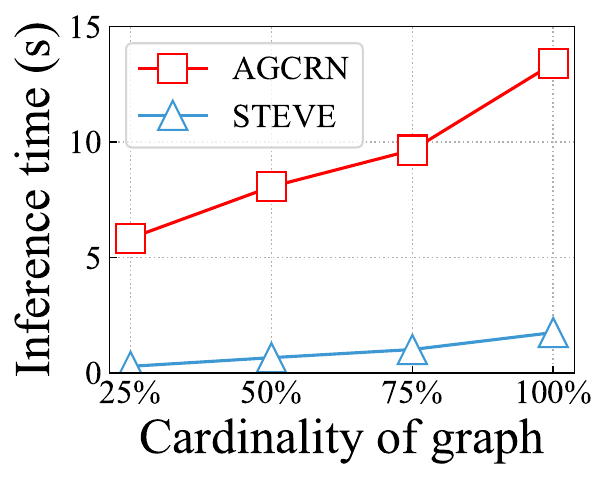}
    }\vspace{-4mm}
    \caption{Scalability performance vs. cardinality}\label{fig:scalability}\vspace{-5mm}
\end{figure}

In the section, we explore the scalability performance of \model compared with AGCRN (the best baseline), focusing on their ability to handle variations in dataset size and graph size. The evaluation employs the BJTaxi dataset that contains traffic data from 1024 graph nodes over 4 months.
\fig~\ref{fig:scalability} depicts the experimental results. Regarding the dataset size, 25\% denotes a one-month dataset, 50\% denotes a two-month dataset, and so on. For the graph size, we decompose the input graph into four connected subgraphs with the same node number. Here, 25\% implies using nodes from the first subgraph to extract an adjacency matrix from the original one, 50\% involves nodes from the first two subgraphs, and so on.
From \fig~\ref{fig:scalability}, we can observe that the prediction time for both models increases as the dataset and graph size scale. However, the increasing trend of AGCRN is sharp, while \model's trend is more stable.
This demonstrates our model's potential scalability in large-scale ST forecasting.

\section{Related Work}\label{sec:related_work}

\textbf{Spatial-Temporal Traffic Forecasting} has received increasing attention due to its pivotal role in intelligent transportation management~\cite{ji2022stden,wang2015predictability,libcity}. Early contributions emerged from the time series community and predominantly utilized the ARIMA family to model traffic data~\cite{arima,svr}. 
However, these methods usually rely on stationary assumptions, leading to limited representation power for traffic data.
Recent advancements have introduced a variety of deep learning techniques that do not rely on stationary assumptions, enabling the capture of complex traffic dependencies more effectively. For instance, methods like recurrent neural networks~\cite{wang2019empowering,dcrnn} and temporal convolutional networks~\cite{gwnet,stgcn,wang2022traffic} are employed to capture temporal dependencies. 
Regarding spatial dependencies, convolutional neural networks~\cite{stresnet,nycbike2} are used for grid-based spatiotemporal data, while graph neural networks~\cite{ji2022stden,shao2022decoupled,jiang2023selfsupervised} and attention mechanism~\cite{astgnn,wang2016traffic,pdformer2023} are explored to incorporate road network information. 
Recently, several studies have investigated the confounder issue, concentrating on invariant relation learning~\cite{zhou2023maintain}, front-door adjustment~\cite{deng2023spatio}, or a combination of both front-door and back-door adjustments~\cite{xia2023deciphering}.
However, these methods rely on predefined discrete confounder values that are often impractical in real-world scenarios. Consequently, they struggle to address continuous and unknown confounders, which is our primary focus.

\textbf{Self-Supervised Learning} aims to distill valuable information from input data to enhance the quality of representations~\cite{ji2022precision}. The fundamental paradigm involves initially augmenting input data and subsequently employing self-supervised tasks to serve as pseudo labels for the purpose of representation learning~\cite{ren2022generative,ji2023spatio}. These tasks are usually infused with domain knowledge to encourage representations to exhibit specific characteristics. This approach has achieved remarkable success within various data such as text data~\cite{bert}, image data~\cite{chen2020simple}, and audio data~\cite{oord2018representation}. Motivated by these works, we devise customized self-supervised tasks tailored to infuse various information into confounder representations.

\textbf{Disentangled Representation Learning} aims to learn identifying and disentangling the underlying factors hidden in the observable data in representation form~\cite{bengio2013representation}, which has been verified to increase the model generality~\cite{disentangle-dvib}.
It was initially used to analyze visual data~\cite{higgins2017betavae} and has recently been introduced to the field of spatiotemporal prediction~\cite{shao2022decoupled}.
Some studies focus on disentangling from the time dimension, \eg seasonal-trend disentanglement and frequency disentanglement~\cite{fang2023stwave, deng2024scnn, woo2022cost}.
Some work focuses on structural disentanglement from the spatial dimension~\cite{deng2021st,ji2023multi,ji2020interpretable}.
However, they are mainly unsupervised disentangling methods, which proved to be unable to disentangle from the corresponding underlying factors~\cite{icml19}.
In contrast, this paper utilizes self-supervised signals to ensure the effectiveness of disentanglement.

\section{Conclusion and Future Work}\label{sec:con}

This paper presented the first attempt to extend back-door adjustment to handle continuous or unknown confounders in deep-learning traffic prediction.
By utilizing a basis vector approach, we proposed a STEVE model that creates a base confounder bank to represent any confounder as an adaptive linear combination of a group of basis confounder representations, with the aid of three self-supervised auxiliary tasks. Then, we decoupled the confounder-irrelevant relations from confounder effects and used both types of relations for robust traffic prediction. 
Extensive experiments over four datasets verified the effectiveness, robustness, and scalability of our model.
In the future, we plan to extract representations of common confounders (such as weather and holidays) to quantify the quantitative impact of these confounders on traffic states and make counterfactual traffic predictions under intervention settings.

\begin{acks}
Prof. Jingyuan Wang's work was partially supported by the National Natural Science Foundation of China (No. 72222022, 72171013, 72242101), and the Special Fund for Health Development Research of Beijing (2024-2G-30121). 
\end{acks}

\bibliographystyle{ACM-Reference-Format}
\bibliography{5-sample}

\appendix

\section{Supplementary Material}

\subsection{Experimental Setting}\label{appx:setting}

\subsubsection{Datasets} We conducted experiments on four commonly used real-world large-scale datasets released by~\cite{ji2023spatio}. These datasets are generated by millions of taxis or bikes on average and contain thousands of time steps and hundreds of regions. The statistical information is in \tab\ref{tab:SoD}. Two of them are bike datasets, while the others are taxi datasets. Bike data record bike rental demands. Taxi data record the number of taxis coming to and departing from a region given a specific time interval, \ie inflow and outflow. 

We give more detailed descriptions of the four datasets as follows. \textbf{NYCTaxi}~\cite{nycbike2} measures the 30-minute level taxi flow from 1/Jan/2015 to 01/Mar/2015. NYCBike series datasets consist of hourly level dataset from 1/Apr/2014 to 30/Sept/2014 (\textbf{NYCBike1}~\cite{stresnet}) and one 30-minute level dataset from 1/Jul/2016 to 29/Aug/2016 (\textbf{NYCBike2}~\cite{nycbike2}).  \textbf{BJTaxi}~\cite{stresnet} is also a 30-minute level taxi dataset from 01/Mar/2015 to 30/Jun/2015, collected in Beijing city.  
For all datasets, the traffic network is constructed by the adjacency relation of regions. 
For a prediction sample at time slot $t$, we use two types of past data as inputs: $i)$ data from 4 hours before $t$, and $ii)$ data from 2 hours before and after the time slots $t-T_{day}$, $t-2T_{day}$, and $t-3T_{day}$, where $T_{day}$ is the number of time slots in one day. The second type incorporates periodicity information into the prediction.
We adopt a sliding window strategy to generate samples, and then split each dataset into the training, validation, and test sets with a ratio of 7:1:2.

\begin{table}[t]
  \centering
  \caption{Statistics of Datasets.}\vspace{-3mm}
  \resizebox{0.8\columnwidth}{!}{
    \setlength{\tabcolsep}{0.8mm}
    \begin{tabular}{rcccc}
    \toprule
    Dataset & NYCTaxi & NYCBike1 & NYCBike2 &  BJTaxi \\
    \midrule
    Time interval & 30 min & 1 hour & 30 min &  30 min \\
    \# regions &  10$\times$20 & 16$\times$8  & 10$\times$20 &  32$\times$32 \\
    \# taxis/bikes & 22m+  & 6.8k+ & 2.6m+ &  34k+ \\
    \# samples & 2880  &4392  & 2880  &  5596 \\
    \bottomrule
    \end{tabular}%
    }\vspace{-5mm}
  \label{tab:SoD}%
\end{table}%

\subsubsection{Baselines} Since traditional statistical models and shallow machine learning methods have proven difficult to effectively model ST traffic data~\cite{agcrn,astgnn}, we compare \model with recent state-of-the-art baselines as follows. 

{\bf \noindent $i)$ Spatial-temporal prediction methods based GNNs:}
\begin{itemize}[leftmargin=*]
    \item \textbf{STGCN}~\cite{stgcn}: a graph convolution-based model that combines 1D-convolution to capture spatial and temporal correlations.
    \item \textbf{AGCRN}~\cite{agcrn}: it enhances the classical graph convolution with an adaptive adjacency matrix and combines it into RNN. 
    \item \textbf{ASTGNN}~\cite{astgnn}: it incorporates self-attention blocks to model the dynamics of traffic data in both temporal and spatial dimensions.
    \item \textbf{HimNet}~\cite{himnet}: it captured spatiotemporal heterogeneity by learning spatial and temporal embeddings, and proposed a novel meta-parameter learning paradigm to learn spatiotemporal-specific parameters from meta-parameter pools.
\end{itemize}

{\bf \noindent $ii)$ Disentanglement-based ST prediction methods:}
\begin{itemize}[leftmargin=*]
    \item \textbf{COST}~\cite{woo2022cost}: a time series model that disentangles seasonal and trend information from a causal lens to enhance model robustness to distribution shifts in time series forecasting.
    \item \textbf{ST-Norm}~\cite{deng2021st}: it introduces temporal and spatial normalization modules to refine the high-frequency and local components of the original ST data, respectively.
    \item \textbf{STWA}~\cite{fang2023stwave}: it disentangles the complex traffic data into stable trends and fluctuating events for accurate prediction.
    \item \textbf{SCNN}~\cite{deng2024scnn}: it disentangles the input data into long-term, seasonal, short-term, and co-evolving components iteratively and then fusing them for spatiotemporal prediction.
\end{itemize}

\begin{figure}[t]
    \centering
    \subfigure[NYCTaxi]{\includegraphics[width=0.24\columnwidth]{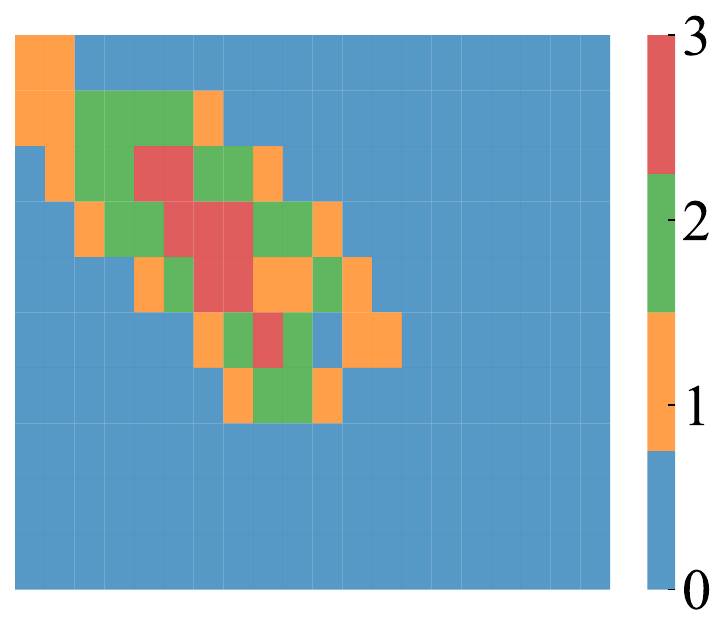}\label{fig:cluster_NYCTaxi}}
    \subfigure[NYCBike1]{\includegraphics[width=0.24\columnwidth]{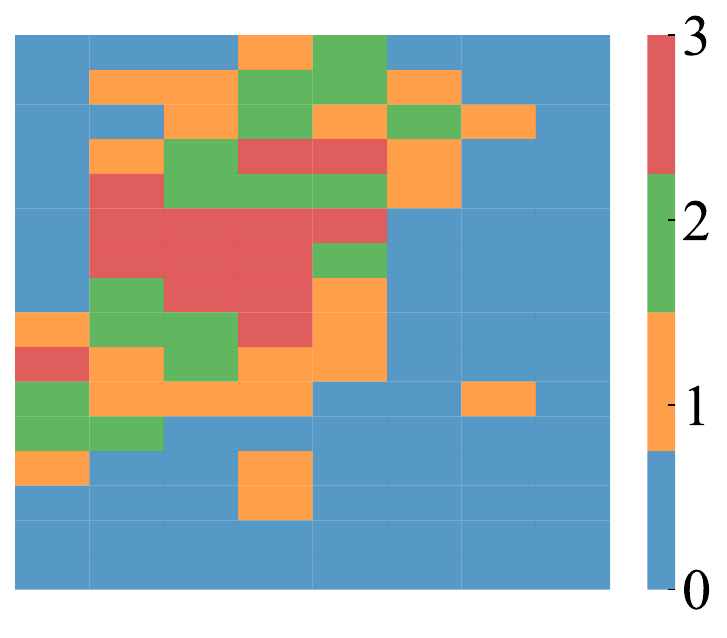}\label{fig:cluster_NYCBike1}}
    \subfigure[NYCBike2]{\includegraphics[width=0.24\columnwidth]{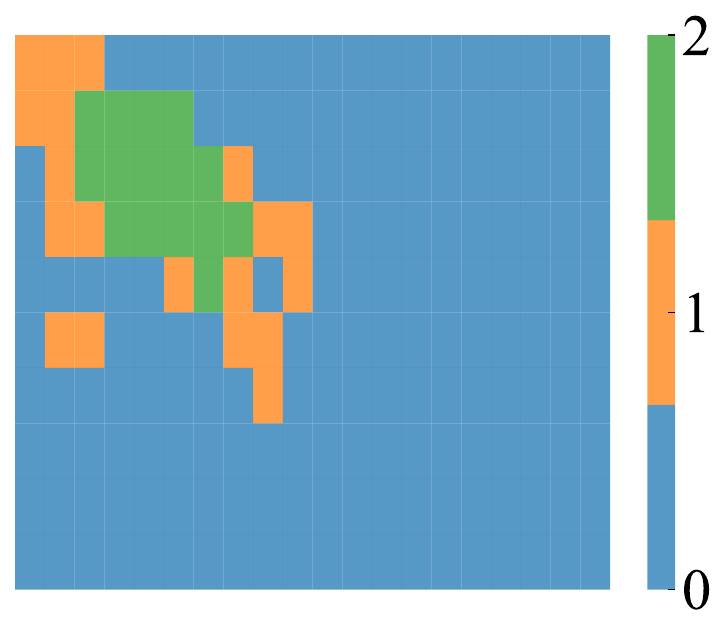}\label{fig:cluster_NYCBike2}}
    \subfigure[BJTaxi]{\includegraphics[width=0.23\columnwidth]{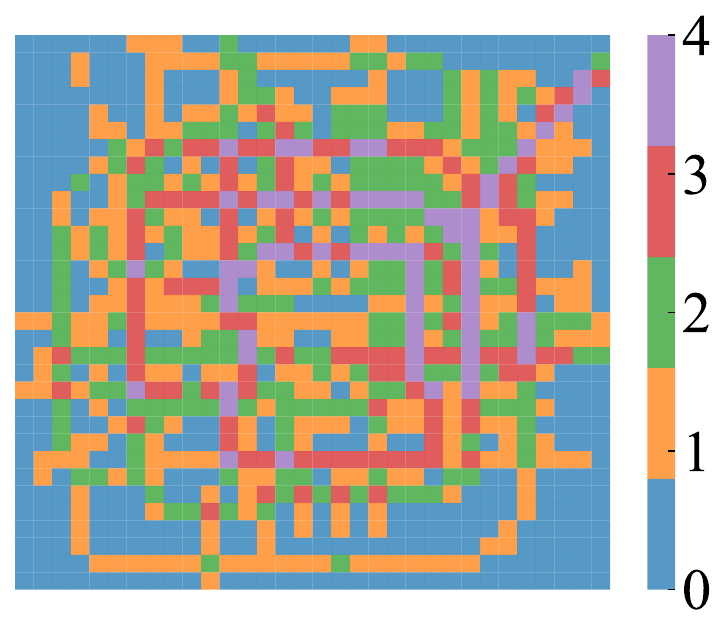}\label{fig:cluster_BJTaxi}}
    \subfigure[Critical Difference (CD) diagram of the Nemenyi test]{\includegraphics[width=0.72\columnwidth]{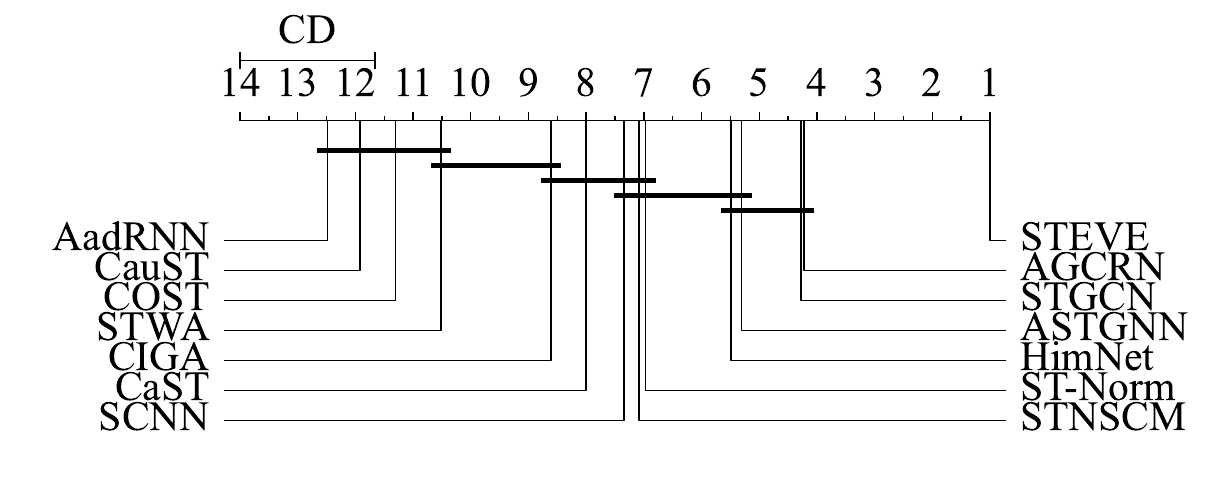}\label{fig:nemenyi}}
    \vspace{-4mm}
    \caption{(a)-(d): Spatial clustering results of all datasets. The cluster identification (ID) is next to the color bar. A larger cluster ID means a higher level of popularity in the corresponding region. (e): CD diagram of the Nemenyi test. The horizontal axis depicts the average ranking of each model across all scenarios of both metrics. Bold black lines connect two models when their ranking difference is below the CD value (at a 5\% significance level), indicating statistical insignificance. Otherwise, they are significantly different.}\label{fig:cluster}\vspace{-5mm}
\end{figure}

\begin{table*}[t]
  \centering
  \caption{Complete performance comparison on NYCBike2 and BJTaxi. The bold/underlined font means the best/the second-best result. Work: Workday. Holi: Holiday. $c_i$: Spatial entity cluster with id $i$. Avg: Average results of different tasks.}\vspace{-3mm}
  \resizebox{0.8\linewidth}{!}{
  \setlength{\tabcolsep}{0.4mm}
  \renewcommand{\arraystretch}{0.9}
        \begin{tabular}{c|c|cc|c|ccc|c|cc|c|ccccc|c}
    \toprule
    \multirow{3}{*}{Model} & Dataset & \multicolumn{7}{c|}{NYCBike2}                         & \multicolumn{9}{c}{BJTaxi} \\
\cline{2-18}          & \multirow{2}{*}{Task} & \multicolumn{3}{c|}{TDS} & \multicolumn{4}{c|}{SDS}      & \multicolumn{3}{c|}{TDS} & \multicolumn{6}{c}{SDS} \\
\cline{3-18}          &       & Work  & Holi  & Avg   & c0    & c1    & c2    & Avg   & Work  & Holi  & Avg   & c0    & c1    & c2    & c3    & c4    & \multicolumn{1}{c}{Avg} \\
    \toprule
    \multirow{2}[1]{*}{STGCN} & MAE   & 5.43  & 5.53  & 5.48  & 3.94  & 4.90  & 7.05  & 5.30  & 12.52  & 11.77  & 12.14  & 4.97  & 9.45  & 13.84  & 20.64  & 30.13  & 15.80  \\
          & MAPE  & 25.09  & 30.71  & 27.90  & 37.47  & 26.47  & 20.17  & 28.04  & 14.91  & 19.34  & 17.13  & 23.88  & 15.41  & 12.38  & \underline{10.56} & 9.34  & 14.31  \\
    \multirow{2}[0]{*}{AGCRN} & MAE   & 5.35  & 5.43  & \underline{5.39} & 3.80  & \underline{4.74} & 7.00  & \underline{5.18} & \underline{11.99} & 11.11  & \underline{11.55} & 5.93  & 11.16  & 20.51  & 21.71  & 31.44  & 18.15  \\
          & MAPE  & 24.62  & 30.15  & 27.39  & 36.73  & 26.04  & \underline{19.75} & \underline{27.51} & 14.68  & 18.98  & 16.83  & 28.87  & 15.37  & 12.77  & 11.46  & 9.29  & 15.55  \\
    \multirow{2}[0]{*}{ASTGNN} & MAE   & \underline{5.25} & 5.62  & 5.43  & 3.58  & 4.94  & 7.34  & 5.29  & 12.09  & \underline{11.03} & 11.56  & \underline{4.95} & \underline{9.37} & \underline{13.40} & \underline{19.74} & \underline{27.89} & \underline{15.07} \\
          & MAPE  & 28.16  & 35.24  & 31.70  & 37.62  & 27.86  & 22.06  & 29.18  & 15.31  & 19.77  & 17.54  & 22.99  & \underline{15.06} & \underline{12.04} & 10.90  & 9.53  & \underline{14.08} \\
    \multirow{2}[1]{*}{HimNet} & MAE   & 5.31  & 5.67  & 5.49  & 3.95  & 4.75  & 7.11  & 5.27  & 12.60  & 11.48  & 12.04  & 5.09  & 9.62  & 14.18  & 20.84  & 30.69  & 16.08  \\
          & MAPE  & \underline{24.49} & 31.44  & 27.96  & 36.50  & \underline{25.87} & 20.61  & 27.66  & \underline{14.45} & 18.98  & \underline{16.72} & \underline{22.49} & 15.42  & 12.59  & 10.62  & 9.43  & 14.11  \\
    \midrule
    \multirow{2}[1]{*}{COST} & MAE   & 7.06  & 7.50  & 7.28  & 3.91  & 5.87  & 10.38  & 6.72  & 14.05  & 13.87  & 13.96  & 5.18  & 10.35  & 15.75  & 24.80  & 37.31  & 18.68  \\
          & MAPE  & 31.23  & 39.32  & 35.28  & 36.68  & 31.92  & 33.54  & 34.05  & 17.10  & 22.41  & 19.76  & 23.76  & 18.62  & 15.91  & 14.50  & 12.94  & 17.15  \\
    \multirow{2}[0]{*}{ST-Norm} & MAE   & 5.57  & \underline{5.39} & 5.48  & 3.46  & 5.04  & 7.01  & 5.17  & 13.26  & 13.36  & 13.31  & 5.79  & 10.71  & 15.50  & 22.19  & 31.08  & 17.05  \\
          & MAPE  & 26.25  & \underline{27.62} & \underline{26.94} & 33.33  & 28.86  & 21.42  & 27.87  & 16.75  & \underline{18.97} & 17.86  & 25.63  & 16.87  & 13.38  & 10.79  & \underline{8.99} & 15.13  \\
    \multirow{2}[0]{*}{STWA} & MAE   & 10.00  & 8.87  & 9.44  & 3.86  & 7.77  & 15.62  & 9.08  & 13.24  & 13.25  & 13.25  & 5.22  & 10.16  & 15.30  & 23.09  & 33.98  & 17.55  \\
          & MAPE  & 36.38  & 51.83  & 44.11  & 34.07  & 44.37  & 44.06  & 40.83  & 15.52  & 21.87  & 18.69  & 23.99  & 17.13  & 14.27  & 12.19  & 10.71  & 15.66  \\
    \multirow{2}[1]{*}{SCNN} & MAE   & 5.78  & 5.65  & 5.71  & 3.64  & 5.15  & 7.82  & 5.54  & 12.68  & 11.80  & 12.24  & 5.25  & 9.82  & 14.39  & 21.16  & 30.05  & 16.13  \\
          & MAPE  & 25.99  & 31.24  & 28.62  & 34.60  & 28.01  & 22.70  & 28.44  & 15.24  & 19.37  & 17.31  & 23.67  & 16.00  & 13.11  & 11.00  & 9.49  & 14.65  \\
    \midrule
    \multirow{2}[1]{*}{AdaRNN} & MAE   & 8.18  & 7.35  & 5.96  & 5.02  & 7.56  & 13.53  & 8.71  & 19.63  & 17.78  & 18.71  & 6.33  & 13.99  & 22.63  & 33.67  & 52.23  & 25.77  \\
          & MAPE  & 36.54  & 28.47  &32.51  & 39.72  & 38.49  & 40.66  & 39.62  & 21.89  & 28.79  & 25.34  & 27.86  & 24.17  & 22.03  & 20.02  & 17.50  & 22.32  \\
    \multirow{2}[0]{*}{CIGA} & MAE   & 6.05  & 5.86  & 5.96  & \underline{3.31} & 5.64  & 9.56  & 6.17  & 13.47  & 12.69  & 13.08  & 7.81  & 12.09  & 16.29  & 22.01  & 31.03  & 17.85  \\
          & MAPE  & 31.49  & 28.45  & 29.97  & 37.79  & 28.77  & 29.34  & 31.97  & 16.71  & 22.24  & 19.48  & 29.14  & 16.88  & 13.91  & 12.99  & 10.58  & 16.70  \\
    \multirow{2}[0]{*}{STNSCM} & MAE   & 6.15  & 5.76  & 5.96  & 6.11  & 6.71  & \underline{6.82} & 6.54  & 13.80  & 11.29  & 12.55  & 6.79  & 10.87  & 14.09  & 21.96  & 29.23  & 16.59  \\
          & MAPE  & 27.88  & 31.13  & 29.51  & \underline{29.68} & 27.62  & 27.68  & 28.33  & 16.96  & 19.36  & 18.16  & 23.87  & 18.43  & 14.22  & 12.22  & 9.64  & 15.67  \\
    \multirow{2}[0]{*}{CauSTG} & MAE   & 7.26  & 5.50  & 6.38  & 4.30  & 6.46  & 10.78  & 7.18  & 17.31  & 25.93  & 21.62  & 6.35  & 13.51  & 22.75  & 37.88  & 58.99  & 27.90  \\
          & MAPE  & 28.87  & 28.72  & 28.80  & 38.22  & 30.75  & 25.85  & 31.61  & 19.27  & 30.15  & 24.71  & 28.44  & 20.77  & 19.69  & 18.95  & 18.07  & 21.18  \\
    \multirow{2}[1]{*}{CaST} & MAE   & 6.25  & 7.19  & 6.72  & 5.01  & 5.70  & 8.59  & 6.43  & 12.93  & 11.76  & 12.35  & 5.28  & 10.11  & 14.42  & 21.12  & 31.25  & 16.44  \\
          & MAPE  & 28.60  & 36.19  & 32.40  & 45.59  & 27.31  & 24.07  & 32.32  & 16.45  & 21.29  & 18.87  & 24.22  & 17.41  & 14.35  & 11.04  & 10.47  & 15.50  \\
    \midrule
    \multirow{2}[2]{*}{STEVE} & MAE   & \textbf{4.75} & \textbf{4.98} & \textbf{4.87} & \textbf{2.82} & \textbf{4.46} & \textbf{6.64} & \textbf{4.64} & \textbf{11.22} & \textbf{10.66} & \textbf{10.94} & \textbf{4.62} & \textbf{8.90} & \textbf{12.94} & \textbf{18.91} & \textbf{27.00} & \textbf{14.47} \\
          & MAPE  & \textbf{20.57} & \textbf{25.31} & \textbf{22.94} & \textbf{24.72} & \textbf{23.61} & \textbf{18.88} & \textbf{22.40} & \textbf{13.97} & \textbf{18.95} & \textbf{16.46} & \textbf{22.19} & \textbf{15.05} & \textbf{12.03} & \textbf{10.11} & \textbf{8.65} & \textbf{13.61} \\
    \bottomrule
    \end{tabular}%
    }\vspace{-4mm}
  \label{tab:maintable}%
\end{table*}%

{\bf \noindent $iii)$ Models considering distribution shift:}
\begin{itemize}[leftmargin=*]
    \item \textbf{AdaRNN}~\cite{adarnn}: a purely sequential model that addresses distribution shift challenges. It clusters historical time sequences into different classes and dynamically matches input data to these classes to identify contextual information.
    \item \textbf{CIGA}~\cite{chen2022learning}: it is a graph model that captures the invariance of graphs via causal models to guarantee generalization under various distribution shifts.
    \item \textbf{STNSCM}~\cite{deng2023spatio}: it neuralizes a structural causal model and incorporates external conditions such as time factors and weather for spatiotemporal traffic prediction in OOD scenarios. For fair comparisons, we only use time factors because weather data is not available in all datasets.
    \item \textbf{CauSTG}~\cite{zhou2023maintain}: it is a spatiotemporal model that captures invariant relations for generalization to distribution shift data.
    \item \textbf{CaST}~\cite{xia2023deciphering}: it leverages a causal lens to handle the temporal distribution shift issue by back-door adjustment and captures the dynamic spatial causation via edge-level graph convolution.
\end{itemize}

\subsubsection{Parameter Setting for STEVE}
We conducted a grid search to optimize the hyperparameters of our model across all datasets, focusing on parameters such as hidden dimension $D$, momentum coefficient $\gamma$, number of base confounders $K$, batch size, kernel sizes in TCL and GCL, and learning rate. The best kernel size is 3 for all datasets. The batch sizes of NYCTaxi and NYCBike2 are 64, while those of NYCBike1 and BJTaxi are 32. The rest of the hyper-parameter settings are in Appendix~\ref{appx:para_sen}.

\subsection{More Experimental Results}
\subsubsection{Performance on NYCBike2 and BJTaxi}\label{appx:main_tab}
\tab\ref{tab:maintable} presents the complete results for the baselines on NYCBike2 and BJTaxi. We can observe similar phenomena as the results on NYCTaxi and NYCBike1, \eg our \model surpasses other baselines in most cases.

\begin{figure}[t]
    \centering
    \subfigure[$\gamma$ : Momentum coefficient]{
    \includegraphics[width=0.9\columnwidth]{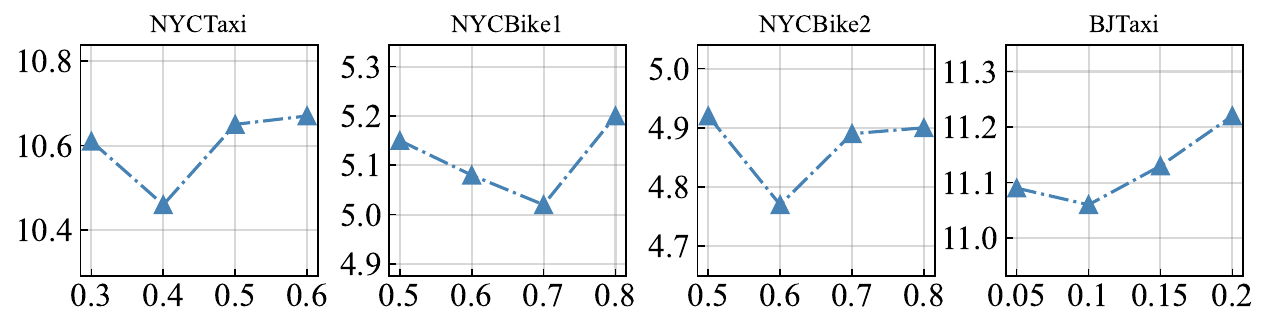}\label{fig:gamma}
    }\vspace{-4mm}
    \subfigure[$K$ : Number of base confounder variables]{
    \includegraphics[width=0.9\columnwidth]{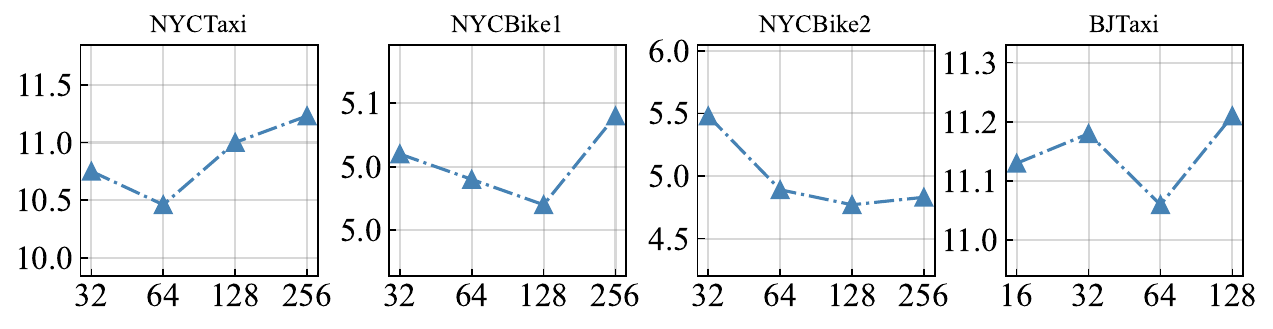}\label{fig:codek}
    }\vspace{-4mm}
    \subfigure[$D$ : Hidden dimension]{
    \includegraphics[width=0.9\columnwidth]{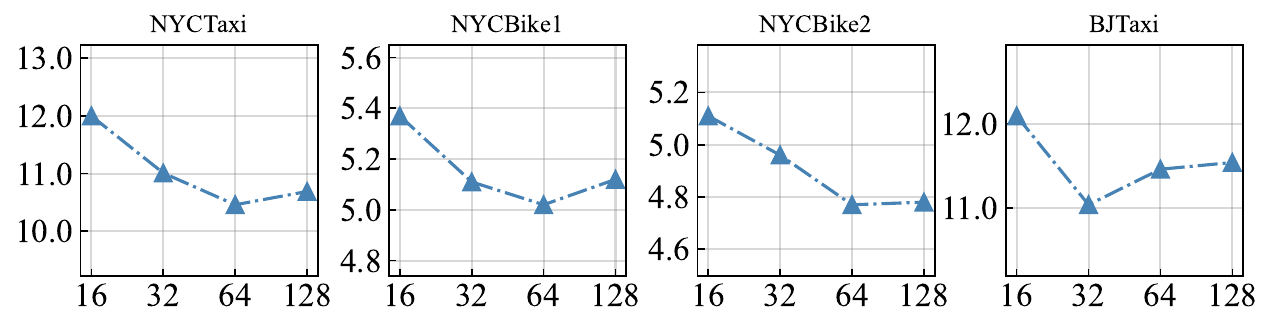}\label{fig:embed}
    }\vspace{-4mm}
    \caption{Parameter sensitivity of \model using MAE metric.}\vspace{-6mm}\label{fig:pm_sen}
\end{figure}
\subsubsection{Spatial Clustering Results}

Recall that in the spatial scenario, we split all regions into clusters to simulate urban functional areas.
Since there is no function label, we use $k$-means clustering algorithm to label the regions. 
The best $k$ is determined by the Silhouette Coefficient metric~\cite{rousseeuw1987silhouettes}. 
The input of $k$-means is $(mean, median, standard~deviation)$ of each region's historical traffic flows. 
\fig~\ref{fig:cluster_NYCBike1}-\ref{fig:cluster_BJTaxi} presents the clustering results of all datasets.
The clustering results exhibit some meaningful patterns, \eg the clusters of the BJTaxi dataset imply the suburbs (ID 0) and ring roads (ID 3). 

\subsubsection{Significance Test}\label{appx:sig_test}
To further emphasize the substantial improvement of our \model over the baseline models, we draw the critical difference (CD) diagram to conduct a Nemenyi significance test. As shown in \fig~\ref{fig:nemenyi}, we can observe that our \model outperforms the best baseline significantly at a 5\% significance level.

\subsubsection{Impact of Hyper-parameters}\label{appx:para_sen}

In this part, we conduct experiments to analyze the impacts of critical hyper-parameters: the momentum coefficient $\gamma$, the number of base confounders $K$, and the hidden dimension $D$, with results in \fig~\ref{fig:pm_sen}.
Firstly, the effect of $\gamma$ is shown in \fig~\ref{fig:gamma}, where we vary it from 0.1 to 0.9 individually and omit some values for better plotting. 
The results indicate that 0.4 is the optimal setting for the NYCTaxi dataset, 0.7 is optimal for the NYCBike1 dataset, 0.6 is optimal for the NYCBike2 dataset, and 0.1 is optimal for the BJTaxi dataset. The variation in optimal settings across datasets is attributed to the distinct impact of confounders.
Secondly, the effect $K$ is shown in  \fig~\ref{fig:codek}. We can observe that a setting of 64 is optimal for the NYCTaxi and BJTaxi datasets, while a setting of 128 is optimal for the NYCBike1 and NYCBike2 datasets. 
Thirdly, the effect of hidden dimension $D$ is given in \fig~\ref{fig:embed}, where we vary it in the set $\{16, 32, 64, 128\}$. The results indicate 64 as the optimal settings for NYCTaxi, NYCBike1, and NYCBike2  datasets and 32 for BJTaxi. Since different datasets have different spatiotemporal dependencies, it is reasonable to use different hidden dimensions for them.

\end{document}